\documentclass[runningheads]{llncs}

\usepackage{eccv}

\usepackage{eccvabbrv}

\usepackage{graphicx}
\usepackage{booktabs}

\usepackage{tabularx}
\usepackage{pifont}
\usepackage{adjustbox}
\newcommand{\cmark}{\ding{51}}%
\newcommand{\xmark}{\ding{55}}%
\newcommand{\ieno}{\textit{i}.\textit{e}.}

\usepackage[accsupp]{axessibility}  

\usepackage{hyperref}

\usepackage{orcidlink}
\usepackage{indentfirst}
\usepackage{tabularx}
\usepackage{multirow}
\usepackage{amssymb}
\usepackage{bm}      
\usepackage{bbding}
\usepackage{array}
\usepackage{color, colortbl}
\definecolor{Gray}{gray}{0.9}
\definecolor{Gray2}{gray}{0.8}

\begin{document}

\title{HIMO: A New Benchmark for Full-Body Human Interacting with Multiple Objects} 

\titlerunning{HIMO: Full-Body Human Interacting with Multiple Objects}

\author{Xintao Lv\inst{1*} \orcidlink{0009-0005-9986-4080} \and
Liang Xu\inst{1,2*}\orcidlink{0000-0002-6441-4443} \and
Yichao Yan\inst{1\dagger}\orcidlink{0000-0003-3209-8965} \and
Xin Jin\inst{2}\orcidlink{0000-0002-1820-8358} \and
Congsheng Xu\inst{1}\orcidlink{0009-0007-2619-8827} \and \\
Shuwen Wu\inst{1}\orcidlink{0009-0002-6601-3253} \and
Yifan Liu\inst{1}\orcidlink{0009-0008-5158-9667} \and
Lincheng Li\inst{3}\orcidlink{0000-0003-3626-4094} \and
Mengxiao Bi\inst{3}\orcidlink{0009-0007-6680-481X} \and \\
Wenjun Zeng\inst{2}\orcidlink{0000-0003-2531-3137} \and
Xiaokang Yang\inst{1}\orcidlink{0000-0003-4848-2304} 
}

\authorrunning{X.~Lv et al.}

\institute{MoE Key Lab of Artificial Intelligence, AI Institute, Shanghai Jiao Tong University, Shanghai, China \and
Ningbo Institute of Digital Twin, Eastern Institute of Technology, Ningbo, China\\
\and
NetEase Fuxi AI Lab\\
{\small\url{https://lvxintao.github.io/himo}}
\vspace{-0.5em}
}

\maketitle

\let\thefootnote\relax\footnotetext{$^*$Equal contribution}
\let\thefootnote\relax\footnotetext{$^\dagger$Corresponding author}

\begin{figure}[h]
    \centering
    \vspace{-0.5cm}
    \includegraphics[trim=0 0 0 0, clip,width=0.9\textwidth]{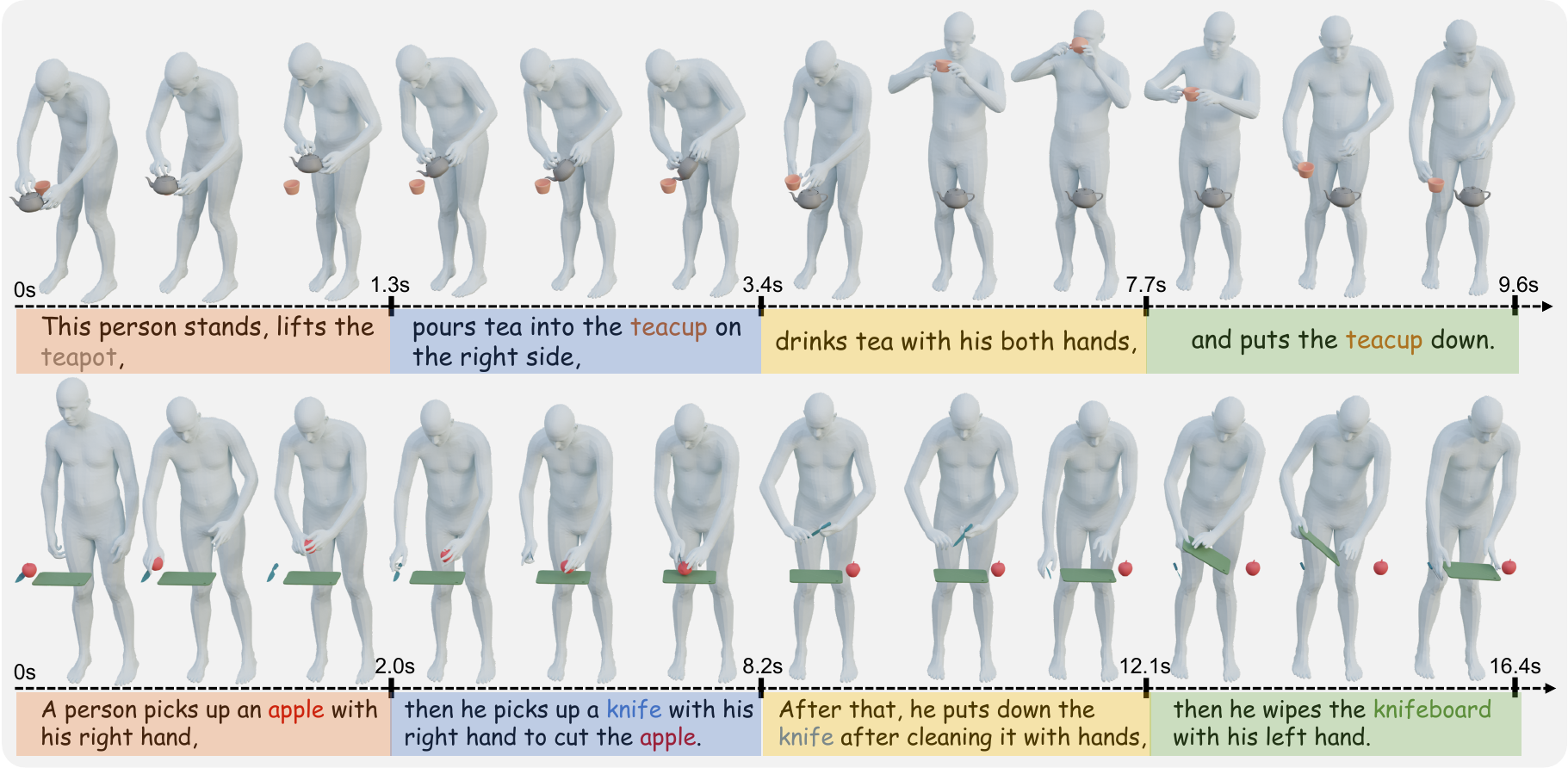}
    \vspace{-0.3cm}
   \caption{\textbf{Overview.} HIMO is a large-scale dataset of full-body interacting with multiple objects, with fine-grained textual descriptions. The long interaction sequences and the corresponding texts are elaborately segmented temporally, facilitating the detailed timeline control of multi-step human-object interactions.}
    \label{fig:teaser}
    \vspace{-1.2cm}
\end{figure}

\begin{abstract}
Generating human-object interactions (HOIs) is critical with the tremendous advances of digital avatars. Existing datasets are typically limited to humans interacting with a single object while neglecting the ubiquitous manipulation of multiple objects.
Thus, we propose \textbf{HIMO}, a large-scale MoCap dataset of full-body human interacting with multiple objects, containing \textbf{3.3K} 4D HOI sequences and \textbf{4.08M} 3D HOI frames. We also annotate \textbf{HIMO} with detailed textual descriptions and temporal segments, benchmarking two novel tasks of HOI synthesis conditioned on either the whole text prompt or the segmented text prompts as fine-grained timeline control. To address these novel tasks, we propose a dual-branch conditional diffusion model with a mutual interaction module for HOI synthesis. Besides, an auto-regressive generation pipeline is also designed to obtain smooth transitions between HOI segments. Experimental results demonstrate the generalization ability to unseen object geometries and temporal compositions.
Our data, codes, and models will be publicly available for research purposes.
\end{abstract}

\section{Introduction}
\label{sec:intro}


Humans constantly interact with objects as daily routines. As a key component for human-centric vision tasks, the ability to synthesize human-object interactions (HOIs) is fundamental with numerous applications in video games, AR/VR, robotics, and embodied AI. 
However, most of the previous datasets and models~\cite{GRAB:2020,bhatnagar22behave,huang2022intercap,fan2023arctic,Jiang_2023_ICCV_chairs} are limited to interacting with a single object, yet neglect the ubiquitous functionality combination of multiple objects.
Intuitively, the multiple objects setting is more practical and allows for broader applications, such as manipulating multiple objects for robotics~\cite{pan2022algorithms,yang2023watch}.

The scarcity of such datasets mainly results in the underdevelopment of the synthesis of human interacting with multiple objects, as listed in~\cref{tab:dataset}.
GRAB~\cite{GRAB:2020} builds the dataset of 4D full-body grasping of daily objects (``4D'' refers to 3D geometry and temporal streams, ``full-body'' emphasizes the body movements and dexterous finger motion), followed by several 4D HOI datasets with RGB modality~\cite{bhatnagar22behave,huang2022intercap}, interacting with articulated~\cite{fan2023arctic} or sittable~\cite{Jiang_2023_ICCV_chairs} objects.
For text-driven HOI generation,~\cite{hoi_diff,cg_hoi} manually label the textual descriptions for the BEHAVE~\cite{bhatnagar22behave} and CHAIRS~\cite{Jiang_2023_ICCV_chairs} datasets, and Li~\etal~\cite{li2023objectmotionguided,li2023controllable} collect a dataset of human interacting with daily objects with textual annotations.
Despite the significant development, all these datasets focus on single-object interactions.
Thus, a dataset of full-body humans interacting with multiple objects incorporated with detailed textual descriptions is highly desired.

Capturing 4D HOIs is challenging due to the subtle finger motions, severe occlusions, and precise tracking of various objects. Compared with a single object, interacting with multiple objects is more complicated regarding the \textit{spatial} movements between human-object and object-object, and the \textit{temporal} schedule of several atomic HOI intervals. For example, the process of a person interacting with a teapot and teacup could be: ``Lift the teapot'' $\to$ ``Pour tea to the teacup'' $\to$ ``Drink tea'', as shown in~\cref{fig:teaser}.
To facilitate the synthesis of \textbf{H}uman \textbf{I}nteracting with \textbf{M}ultiple \textbf{O}bjects, we build a large-scale dataset called \textbf{HIMO}.
We adopt the optical MoCap system to obtain precise body movements and track the motion of objects attached by reflective markers. For the dexterous finger motions, we employ wearable inertial gloves to avoid occlusions.
In total, \textbf{3.3K} 4D HOI sequences with \textbf{34} subjects performing the combinations of \textbf{53} daily objects are presented, resulting in \textbf{4.08M} 3D HOI frames.
Various subjects, object combinations, and interaction patterns also ensure the diversity of \textbf{HIMO}.

To facilitate the study of text-driven HOI synthesis~\cite{tevet2023mdm,Guo_2022_CVPR_humanml,chen2023mldm,Zhang_2023_CVPR_t2mgpt,cg_hoi}, we annotate the HOI sequences with fine-grained textual descriptions.
Different from existing datasets, we additionally segment the long interaction sequences temporally and align the corresponding texts semantically, which allows for fine-grained timeline control, and a flexible schedule of multiple atomic HOI sequences. 
Revisiting the example of pouring water, after training based on our datasets, \textit{novel} HOI compositions such as ``Drink water'' $\to$ ``Lift the teapot'' $\to$ ``Add more tea to the teacup'' could be synthesized benefited from our fine-grained temporal segments.
Although temporal segmentation is widely adopted in video action domain~\cite{ding2023temporal,liu2023reducing,shi2023tridet,li2024otas}, we are the first to introduce the annotations to 4D HOIs. 

\begin{table*}[t]
\caption{\textbf{Dataset comparisons.} We compare the HIMO dataset with existing human-object interaction datasets. ``Multi-object'', ``Segment'' and ``\#'' mean human interacting with multiple objects, temporal segmentation, and the number of.}
\vspace{-0.2cm}
\label{tab:dataset}
\begin{adjustbox}{width=\columnwidth, center}
\begin{tabular}{l|cccc|cccc}
\hline
Dataset &  \multicolumn{4}{c}{\textbf{Modality}}  & \multicolumn{4}{c}{\textbf{Scale}} \\
& Hand & Multi-object & Text & Segment & \#Subject & \#Object & \#Seq & \#Frame \\ \hline
GRAB~\cite{GRAB:2020} & \cmark & \xmark & \xmark & \xmark & 10 & 51 & 1,334 & - \\
BEHAVE~\cite{bhatnagar22behave} & \xmark & \xmark & \xmark & \xmark & 8 & 20 & 321 & 15K \\
InterCap~\cite{huang2022intercap} & \xmark & \xmark & \xmark & \xmark & 10 & 10 & 223 & 67K \\
ARCTIC~\cite{fan2023arctic} & \cmark & \xmark & \xmark & \xmark & 9 & 11 & 339 & 2.1M \\
CHAIRS~\cite{Jiang_2023_ICCV_chairs} & \cmark & \xmark & \xmark & \xmark & 46 & 81 & 1,390 & - \\
Li \etal~\cite{li2023objectmotionguided} & \cmark & \xmark & \cmark & \xmark & 17 & 15 & - & - \\
\hline
\cellcolor{Gray}\textbf{HIMO(Ours)} & \cellcolor{Gray}\cmark & \cellcolor{Gray}\cmark & \cellcolor{Gray}\cmark & \cellcolor{Gray}\cmark & \cellcolor{Gray}34 & \cellcolor{Gray}53 & \cellcolor{Gray}3,376 & \cellcolor{Gray}4.08M \\ 
\hline
\end{tabular}
\end{adjustbox}
\vspace{-0.2cm}
\end{table*}

Our HIMO dataset enables two novel generative tasks: 1) Text-driven HOI synthesis with multiple objects (denoted as \texttt{HIMO-Gen}); 2) \texttt{HIMO-Gen} conditioned on segmented texts as timeline control (denoted as \texttt{HIMO-SegGen}).
Concretely, instead of single text as the condition for \texttt{HIMO-Gen}, \texttt{HIMO-SegGen} requires a series of multiple consecutive texts as guidance.
For \texttt{HIMO-Gen}, we generate the synchronized human motion and object motions conditioned on the language descriptions and the initial states of the human and the objects.
Inspired by previous works on modeling human-human interaction generation~\cite{priormdm,intergen,xu2024regennet,xu2023inter} and human interacting with single object~\cite{wu2022saga,li2023controllable,xu2023interdiff,hoi_diff,li2023objectmotionguided}, it is straightforward to design a dual-branch conditional diffusion model to generate the human and object motion, respectively. 
However, this naive solution suffers from the following challenges.
First, the generated human motion and object motions could be spatio-temporally misaligned. Second, implausible contacts between human-object and object-object are commonly encountered. To guarantee the coordination between human motion and object motions, we further design a mutual interaction module implemented as a stack of multiple mutual-attention layers. We also add an object-pairwise loss based on the relative distance between the geometry of objects to generate plausible object movements. Experiments show that our method can synthesize realistic and coordinated HOIs.

Furthermore, for the task of \texttt{HIMO-SegGen}, each segmented HOI clip paired with the corresponding text can be viewed as a sample to train the \texttt{HIMO-Gen} model, while the key of \texttt{HIMO-SegGen} lies in the smoothness between the generated clips.
We introduce a simple yet effective scheme to auto-regressively generate one HOI clip at a time, conditioning the next clip generation with the last few frames of the previous clip.
Experiment results show that conditioning on 10 frames achieves the best performance.
Besides, we also show the generalization ability of our model to unseen objects and novel HOI compositions.
Overall, our contributions can be summarized as follows:
\begin{itemize}
\item We collect a large-scale dataset of full-body human interacting with multiple daily objects called \textbf{HIMO}, with precise and diverse 4D-HOI sequences.
\item We annotate the detailed textual descriptions for each long HOI sequence, together with the temporal segmentation labels, facilitating two novel downstream tasks of \texttt{HIMO-Gen} and \texttt{HIMO-SegGen}.
\item We propose a dual-branch conditional diffusion model with a mutual interaction module to fuse the human and object features. We also propose a pipeline to auto-regressively generate the composite HOI sequences.
\end{itemize}

\section{Related Work}
\subsection{Human-Object Interaction Datasets}

Many human-object interactions together with the hand-object interaction datasets have been proposed to facilitate the development of modeling HOIs.
Several works~\cite{Liu_2022_CVPR_hoi4d,Jian_2023_ICCV_affordpose,Hampali_2022_CVPR_Kypt_Trans_h2o3d,hampali2020honnotate,Brahmbhatt_2020_ECCV_contactpose,liu2024taco} focusing on hand-object interaction have been proposed. Though effective, modeling hand-object interactions falls significantly short of comprehending the nature of human-object interaction, since the interaction process involves more than just hand participation.
Full-body interactions~\cite{GRAB:2020,bhatnagar22behave,huang2022intercap,Jiang_2023_ICCV_chairs,fan2023arctic,xu2021d3dhoi,kitwholebody,zhang2023neuraldome,zhang2022couch} may enhance our understanding of HOIs. 
GRAB~\cite{GRAB:2020}, InterCap~\cite{huang2022intercap} and ARCTIC~\cite{fan2023arctic} broaden the range of 4D-HOIs to full-body interactions. 
However, all of these full-body datasets only involve one single object, while the combination of several different objects is common in our daily lives. 
The KIT Whole-body dataset~\cite{kitwholebody} contains motion in which subjects manipulate several objects simultaneously. However, the human body is represented as a humanoid model instead of a more realistic SMPL-X model. 
We address the deficiency by introducing a full-body 4D-HOI dataset that involves full-body interaction with multiple objects. Comparisons with the existing HOI datasets are listed in~\cref{tab:dataset}.

\subsection{Text-driven Motion Generation}

Text-driven human motion generation aims at generating human motions based on textual descriptions. Benefiting from large-scale motion-text dataset like KIT-ML~\cite{plappert2016kit}, BABEL~\cite{babel} and HumanML3D~\cite{Guo_2022_CVPR_humanml}, numerous works~\cite{petrovich22temos,tevet2023mdm,chen2023mldm,Zhang_2023_CVPR_t2mgpt,Zhou_2023_CVPR_ude,dabral2022mofusion,li2023objectmotionguided,xu2023actformer} have emerged. 
TEMOS~\cite{petrovich22temos} proposes a transformer-based VAE to encode motion and an additional DistillBert~\cite{sanh2020distilbert} to encode texts. 
MDM~\cite{tevet2023mdm} adopts the diffusion model~\cite{ho2020ddpm} for motion generation.
MLDM~\cite{chen2023mldm} denoises the low-dimensional latent code of motion instead of the raw motion sequences. 
T2M-GPT~\cite{Zhang_2023_CVPR_t2mgpt} utilizes VQ-VAE~\cite{oord2018vqvae} to compress motion sequences into a learned codebook and then generate human motion in an auto-regressive fashion. GMD~\cite{karunratanakul2023gmd} further adds keyframe conditions to synthesize controllable human motion.
In this work, we base our model on the widely adopted MDM~\cite{tevet2023mdm} framework.

\subsection{Human-Object Interaction Generation}

Generating plausible and realistic human-object interaction sequences has attracted much attention in recent years~\cite{GRAB:2020,Taheri_2022_CVPR_goal,tendulkar2022flex,manipnet,Zheng_2023_CVPR_cams,wu2022saga,xu2023interdiff,ghosh2022imos,taheri2023grip,braun2023physically,christen2022dgrasp,li2023taskoriented,kulkarni2023nifty,li2023objectmotionguided,hoi_diff,cg_hoi}. 
GRAB~\cite{GRAB:2020} generates plausible hand poses grasping a given object. 
GOAL~\cite{Taheri_2022_CVPR_goal} and SAGA~\cite{wu2022saga} extend it to full-body grasping motion generation.
However, these methods convert the task into the generation of the final pose of a human interacting with the \textit{static} object.
Recent works like InterDiff~\cite{xu2023interdiff} tackle the problem of synthesizing comprehensive full-body interactions with dynamic objects, which leverages a diffusion model~\cite{ho2020ddpm} to predict future HOI with realistic contacts. 
IMoS~\cite{ghosh2022imos} synthesizes vivid HOIs conditioned on an intent label, such as: ``use'', ``lift'' and ``pass''. 
Notice that none of the aforementioned methods can generate full-body HOI involving multiple objects.

\subsection{Temporal Action Composition}

Human motion composition~\cite{athanasiou2022teach,lee2023multiact,li2023sequential,qian2023breaking,zhang2020perpetual,priormdm,petrovich2024multi,barquero2024seamless} with text timeline control aims to generate arbitrary long human motion sequences with smooth and realistic transitions, which can be separated as two categories. First, several auto-regressive methods~\cite{athanasiou2022teach,lee2023multiact,li2023sequential,qian2023breaking,zhang2020perpetual} are developed relying on the availability of motion transition annotations. These methods iteratively generate the subsequent motion based on the already generated motion.
Second, a few diffusion-based methods~\cite{priormdm,zhang2023diffcollage,bar2023multidiffusion,petrovich2024multi,barquero2024seamless} are proposed to modify the sampling process of diffusion models with overlaps of a predefined transition duration.
In our setting of \texttt{HIMO-SegGen}, we also need to generate composite HOI sequences with smooth transitions.
Thus, we adopt a simple yet effective auto-regressive pipeline to iteratively synthesize the HOI sequences.

\section{The HIMO Dataset}

We present HIMO, a large-scale 4D-HOI dataset of full-body human interacting with multiple objects, comprising accurate and diverse full-body human motion, object motions and mutual contacts. Next, we will introduce the data acquisition process in~\cref{sec:data_acquire} and then describe the SMPL-X fitting process and the annotation of textual descriptions and the temporal segments in~\cref{sec:data_annotation}.

\begin{figure}[tb]
  \centering
  \includegraphics[width=1\textwidth]{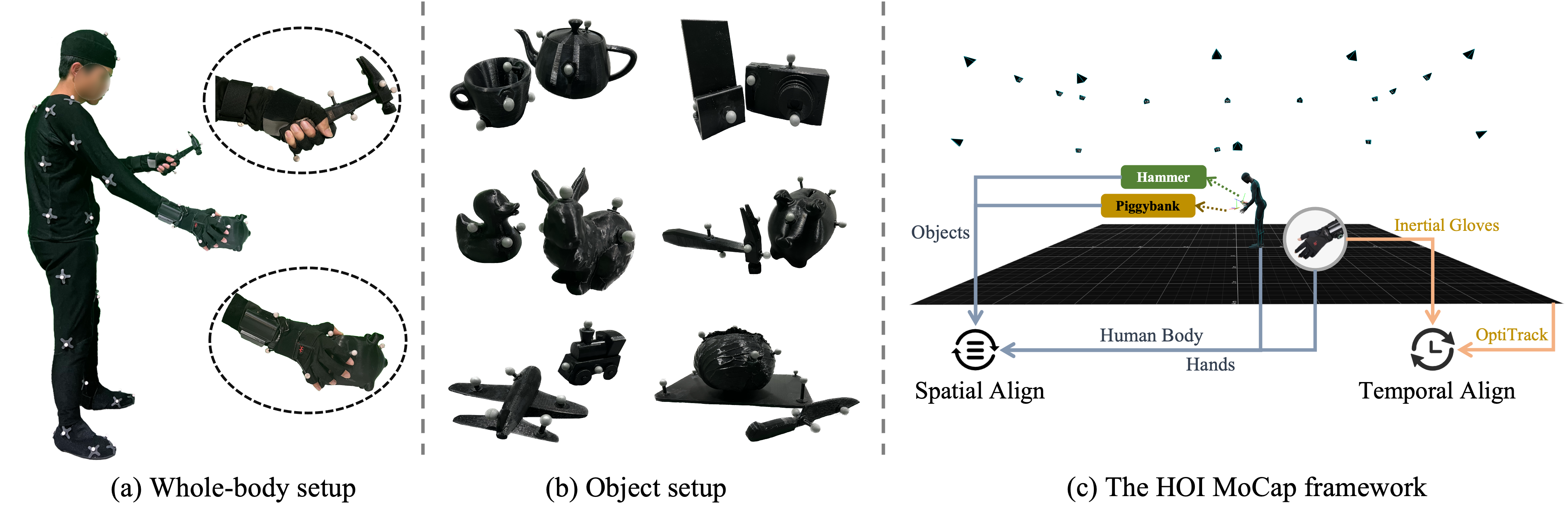}
  \vspace{-0.2cm}
  \caption{\textbf{Overview of the HIMO capture system.} (a). We combine the Optitrack MoCap framework and the inertial gloves as a hybrid MoCap system. (b). Some examples of the 3D printed objects and the attached reflective markers. (c). The whole HOI MoCap framework, where the objects, the human body and hands of the subject are spatially aligned, and the Optitrack and inertial gloves are temporally synchronized.}
  \label{fig:setup}
  \vspace{-0.3cm}
\end{figure}

\subsection{Data Acquisition}
\label{sec:data_acquire}

\noindent\textbf{Hybrid Capture System.} 
To obtain the accurate movement of the human body against occlusions, we follow \cite{GRAB:2020} to adopt the optical MoCap scheme instead of multi-view RGB camera system~\cite{bhatnagar22behave,huang2022intercap}, which guarantees much lower error~\cite{GRAB:2020,xu2023inter}. We believe that high-quality moving sequences are more important than natural images for HOI motion data. An overview of our capturing setup can be seen in~\cref{fig:setup}(a,c).
We deploy Optitrack~\cite{optitrack} MoCap system with 20 PrimeX-22 infrared cameras to capture the body motion, where each subject wearing MoCap suits with 41 reflective markers. Each camera can capture 2K frames at 120fps.
To capture the dexterous finger movements, we select the inertial Noitom Perception Neuron Studio (PNS) gloves~\cite{noitom}, which can precisely record the subtle finger poses in real-time. The inertial gloves can still work well under severe occlusions of human-object and object-object. We frequently re-calibrate the PNS gloves to ensure the capture quality.
Spatially, the locating boards attached to the back of the hands provide the rotation information of the wrists, thus the human body movement can be integrated with the finger movements. Temporally, we utilize the Tentacle Sync device to provide timecodes for Optitrack and PNS gloves to synchronize them.
Apart from the hybrid MoCap system for full-body motion capture, we also set up a Kinect camera to record the RGB videos from the front view with the resolution of 1080$\times$768.

\noindent\textbf{Capturing Objects.} 
We choose 53 common household objects in various scenes including the dining rooms, kitchens, living rooms and studies from ContactDB~\cite{Brahmbhatt_2019_CVPR_contactdb} and Sketchfab~\cite{sketchfab}. 
We further reshape them into appropriate sizes and 3D print them, so that the 3D object geometry is aligned with the real printed objects.
To precisely track the object motion, we attach several (3-6) reflective markers on their surface with strong glue, as shown in~\cref{fig:setup}(b), to track the rigid body composed of the attached markers. The coordinate system of the subject and the objects are naturally aligned. One critical rule of attaching the markers to objects is to mitigate the side effects of manipulating the objects. Empirical results show that 12.5 mm diameter spherical markers achieve more robust tracking performance than smaller markers. Since the Optitrack system tracks the 6-DoF poses of the centroid of the attached markers rather than the centroid of the object, thus we employ a post-calibration process to compensate for the bias between them.
The object category definition is listed in the supplementary.

\noindent\textbf{Recording Process.} Each subject is asked to manipulate the pre-defined combinations of 2 or 3 objects, such as ``pour tea from a \textit{teapot} to a \textit{teacup}" and ``cut an \textit{apple} on a \textit{knifeboard} with a \textit{knife}". Initially, the objects are randomly placed on a table with the height of 74 cm and the subject keeps the rest pose. For each sequence, the subject is required to involve all the provided objects into the manipulation and perform 3 times of each interaction task with different operating patterns for variability. Casual operations yet may not be relevant to the prescribed action are welcome to enhance the complexity of each interaction sequence, such as ``shaking the goblet before drinking from it'' or ``cleaning the knifeboard before cutting food on it''. 
Finally, 34 subjects participated in the collection of HIMO, resulting in \textbf{3,376} HOI sequences, \textbf{9.44} hours and \textbf{4.08M} frames, where 2,486 of them interacting with 2 objects and the other 890 sequences interacting with 3 objects.
More details of the setting of the interaction categories can be found in the supplementary materials.

\subsection{Data Postprocessing}
\label{sec:data_annotation}

\noindent\textbf{Fitting SMPL-X Parameters.} 
The expressive SMPL-X~\cite{SMPL-X:2019} parametric model is widely adopted by recent works~\cite{PIXIE:2021,pymafx2023,lin2023osx,lu2023humantomato,petrovich2024multi} for its generality and flexible body part editing ability.
We also adopt the SMPL-X model to represent the human body with finger articulations.
Formally, the SMPL-X parameters consist of global orient $g\in \mathbb{R}^3$, body pose $\theta_b \in \mathbb{R}^{21\times 3}$, finger poses $\theta_h \in \mathbb{R}^{30\times 3}$, root translation $t\in \mathbb{R}^3$ and shape parameter $\beta\in \mathbb{R}^{10}$. 
An optimization algorithm is adopted to obtain the SMPL-X parameters for each HOI sequence based on our full-body MoCap data. We initialize the subjects' shape $\beta$ based on their height and weight following~\cite{virtual_caliper}. The joint energy term $\mathbb{E}_j$ optimizes body joints to our MoCap data as:
\begin{equation}
    \mathbb{E}_j=\sum\limits_{n=0}\limits^N \sum\limits_{j=0}\limits^{J} \| P_n^j -\hat{P}_n^j \|_2^2,
\end{equation}
where $N$ is the frame number of the sequence, $J$ is the number of human joints, $P_n^j$ and $\hat{P}_n^j$ represent our MoCap joint position and the joint position regressed from the SMPL-X model, respectively. The smoothing term $\mathbb{E}_s$ smooths the motion between frames and mitigates pose jittering as:
\begin{equation}
    \mathbb{E}_s=\sum\limits_{n=0}\limits^{N-1} \sum\limits_{j=0}\limits^{J} \| \hat{P}_{n+1}^j -\hat{P}_n^j \|_2^2.
\end{equation}
Finally, a regularization term $\mathbb{E}_r$ is adopted to regularize the SMPL-X pose parameters from deviating as:
\begin{equation}
    \mathbb{E}_r=\|\theta_b\|_2^2+\|\theta_h\|_2^2.
\end{equation}
In total, the whole optimization objective can be summed as follows:
\begin{equation}
    \mathbb{E}=\alpha\mathbb{E}_j+\lambda\mathbb{E}_s+\gamma\mathbb{E}_r,
\end{equation}
where $\alpha=1,\lambda=0.1,\gamma=0.01$.

\begin{figure}[tb]
  \centering
  \includegraphics[width=1\textwidth]{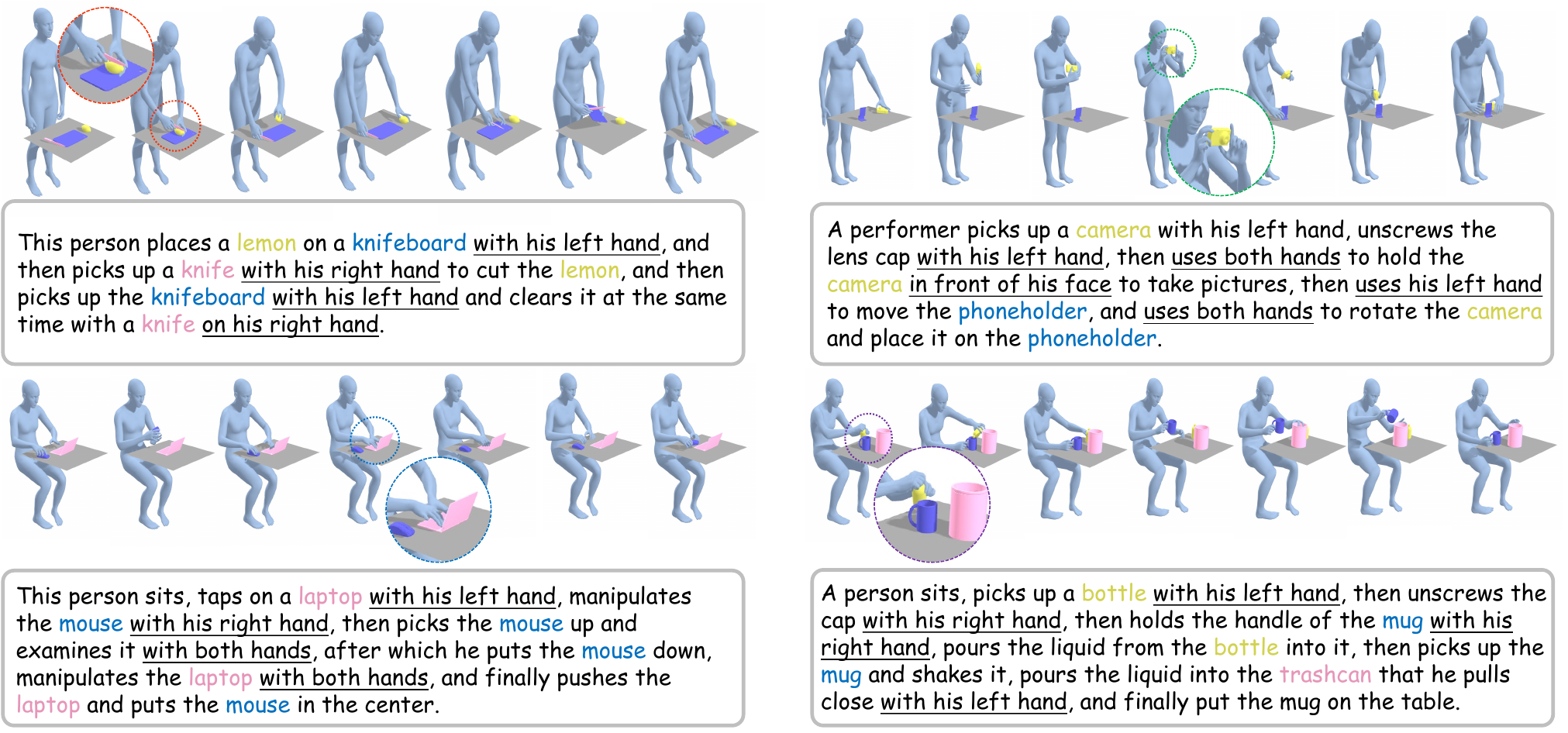}
  \caption{\textbf{More samples of the \textbf{HIMO} dataset.} The subject is asked to interact with 2 or 3 objects based on their functionalities. The textual annotations comprise the detailed interaction order, the interaction pattern and the involved human body parts. Object names are depicted as the corresponding colors and the details are \underline{underlined}.}
  \label{fig:dataset_vis}
\end{figure}

\noindent\textbf{Textual Annotations.} 
Text-driven motion generation thrives greatly thanks to the emergence of text-motion datasets~\cite{plappert2016kit,Guo_2022_CVPR_humanml,wang2022humanise}. To empower the research on text-driven HOI synthesis with multiple objects, \ieno, \texttt{HIMO-Gen}, we elaborately annotate fine-grained textual descriptions of each motion sequence. Different from simple HOI manipulations with a single object and a single prompt, our HIMO dataset includes complex transitions among objects and temporal arrangements among different objects.
Thus, the annotators are asked to describe the entire interaction procedure, emphasizing the detailed manipulation order, \ieno, ``First$\dots$, then$\dots$, finally$\dots$'', the interaction pattern, \ieno, ``picking up'', ``rotating'' and ``pouring'', and the involved human body parts, \ieno, the left hand or right hand.
We exploit an annotation tool based on~\cite{ait-viewer} to enable the annotators to rotate and resize the view to observe the subtle HOIs patterns. 
More visualization results of the text-HOI pairs are presented in~\cref{fig:dataset_vis}.

\noindent\textbf{Temporal Segment Annotations.}
As aforementioned, the temporal segments of the long HOI sequences allow for fine-grained timeline control of decomposing the complex operation process into several simple ones. 
The granularity of the sequence splitting is small, as shown in~\cref{fig:teaser}. For example, the sequence of ``pouring tea from the teapot to the teacup'' can be split into ``lifting the teapot'', ``pouring tea into the teacup'' and ``putting the teacup down''.
Similarly, we implement an annotation tool by~\cite{ait-viewer} to better view the interaction details. The HOI sequence and its corresponding texts are simultaneously split into equal segments.
Statistically, each sequence contains 2.60 segments on average.

\begin{figure}[tb]
  \centering
  \includegraphics[width=1\textwidth]{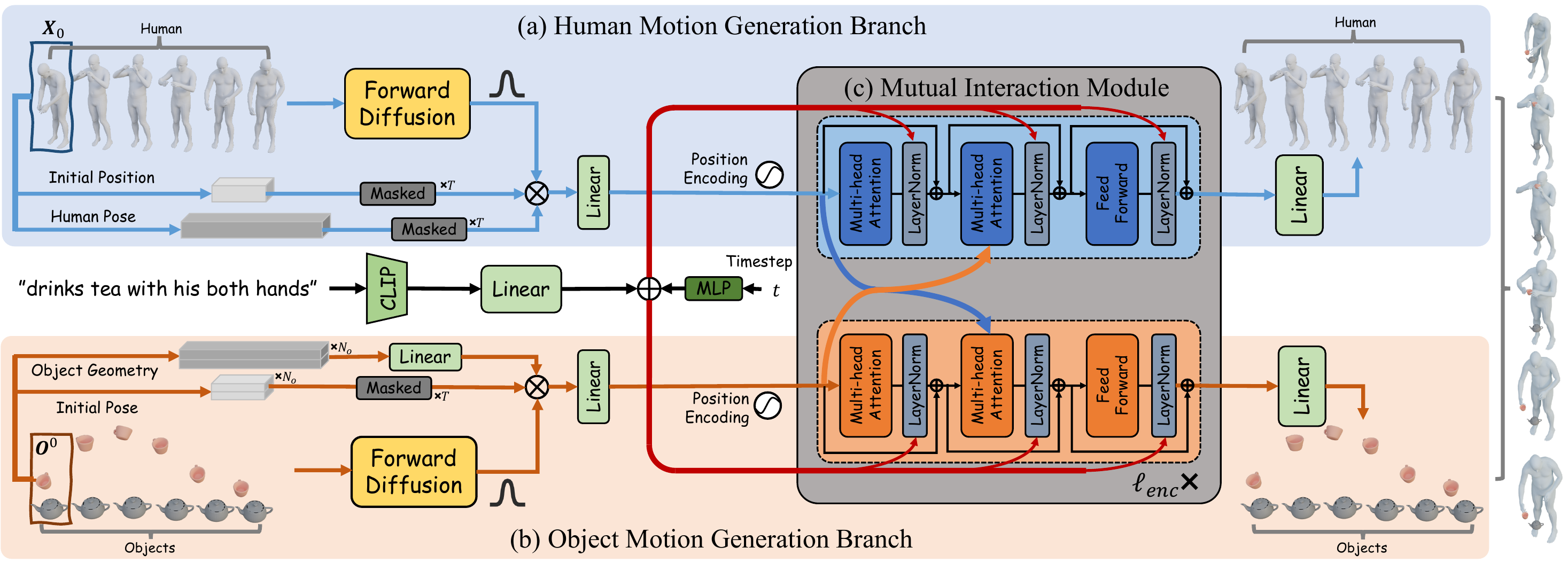}
  \caption{\textbf{Overview of the HOI synthesis framework.} We propose a dual-branch conditional diffusion model to generate the human motion and object motions, respectively, conditioned on the textual description and the initial states of the human and objects. A mutual interaction module is also integrated for information fusion to generate coordinated HOI results.}
  \label{fig:network}
\end{figure}

\section{Text-driven HOI Synthesis}

In this section, we benchmark two \textit{novel} tasks of \texttt{HIMO-Gen} in~\cref{sec:himo_gen} and \texttt{HIMO-SegGen} in~\cref{sec:himo_seggen}. For \texttt{HIMO-Gen}, we propose a dual-branch diffusion-based framework to synthesize the motion of human and objects separately, with meticulously designed mutual interaction module and optimization objectives. For \texttt{HIMO-SegGen}, we apply an auto-regressive generation scheme.

\subsection{The \texttt{HIMO-Gen} Framework}
\label{sec:himo_gen}

\noindent{\textbf{Data Representation.}}
We denote the human motion as $\bm{H}\in \mathbb{R}^{T\times D_h}$ and the multiple object motions as $\bm{O} = \{\bm{O}_i\}_{i=0}^{N_o} \in \mathbb{R}^{T\times D_o}$, where $T$ represents the frame number of the HOI sequence, $N_o$ is the number of the objects, $D_h$ and $D_o$ represent the dimension of the human and object motion, respectively. We adopt the SMPL-X~\cite{SMPL-X:2019} parametric model to represent the human movements.
The pose state of human at frame $i$ is denoted as $\bm{H}^i$, which consists of the global joint positions $P^i\in\mathbb{R}^{52\times3}$, global joint rotation $Q^i\in\mathbb{R}^{52\times6}$ represented by the continuous 6D rotation format~\cite{rot_6d} and translation $t^i\in\mathbb{R}^{3}$. 
We also denote the object motion at frame $i$ as $\bm{O}^i$ (We omit the subscript of the $i$-th object/geometry for better expression), which comprises of the relative rotation $\mathbf{R}^i\in\mathbb{R}^6$ with respect to the input object's frame and its global translation $\mathbf{T}^i\in\mathbb{R}^3$.
To encode the object geometries $\bm{G}=\{\bm{G}_i\}_{i=0}^{N_o}$, we follow prior works~\cite{GRAB:2020,li2023controllable} to adopt the Basis Point Set (BPS) representation~\cite{prokudin2019bps}. We sample 1,024 points from the surface of the object meshes, and then for each sampled point, we calculate the directional vector with its nearest neighbor from the basis points set, resulting in $\bm{G}\in\mathbb{R}^{1024\times3}$ for each object.

\noindent{\textbf{Problem Formulation.}}
Given the textual description $\bm{L}$, the initial states of $\bm{H}^0$ and $\bm{O}^0$ as the \textit{first} frame human motion and object motions and the object geometries $\bm{G}$. Our model aims to generate the spatio-temporally aligned human motion $\bm{H}$ together with the object motions $\bm{O}$ with plausible contacts.

\noindent{\textbf{Dual-branch Diffusion Models.}}
Our proposed framework is demonstrated in~\cref{fig:network}, which consists of two parallel diffusion-based branches for human and object motion generation and a mutual interaction module for feature fusion.
Given a textual description $\bm{L}$ like ``drinks tea with his both hands'', we utilize the CLIP~\cite{radford2021clip} as text encoder followed by a linear layer to obtain the semantic embeddings and then concatenate them with the features of the sampling timestep $t$ extracted by an MLP process.
Take the human motion generation branch as an example, the initial state $\bm{H}^0$ including the initial position and the human pose serves as the conditions. Following~\cite{li2023controllable}, we adopt the masked representation by zero padding $\bm{H}_0$ to $T$ frames.
The masked representations are then concatenated with the noised human motion representation $\bm{H}_t$, where $t$ is the timestep of the forward diffusion process.
For the object motion generation branch, the object geometry $\bm{G}$ is embedded by a linear layer and concatenated with the masked representation of the initial pose of objects and the noised representation of object motions.
We separately feed the obtained human and object representations into a linear layer to obtain the motion embeddings and then pass them into the denoising mutual interaction module with positional encodings.
After that, two linear layers are adopted to obtain the denoised motion representations of human and objects, respectively.

\noindent{\textbf{Mutual Interaction Module.}}
To model the interaction between human and objects, we fuse the features of the two branches via a mutual interaction module as depicted in~\cref{fig:network}(c). The two branches share the same Transformer architecture with $\ell_{enc}$ blocks without sharing weights. 
Each Transformer block is comprised of two attention layers and one feed-forward layer. The first self-attention layer embeds the aggregated human/object features $\bm{H}^{(i)}$ and $\bm{O}^{(i)}$ into embeddings $\mathbf{e}_h^{(i)}$ or $\mathbf{e}_o^{(i)}$.
The second attention layer functions as a \textit{mutual attention} layer, where the key-value pair of the human branch Transformer is provided by the object hidden embedding $\mathbf{e}_o^{(i)}$, and vice versa, which can be formulated as:
\begin{equation}
    \begin{aligned}
        \bm{H}^{(i+1)} = FF(\textit{softmax}(\frac{\mathbf{Q}_h\mathbf{K}_o^{T}}{\sqrt{C}}\mathbf{V}_o))&;
        \bm{O}^{(i+1)}=FF(\textit{softmax}(\frac{\mathbf{Q}_o\mathbf{K}_h^{T}}{\sqrt{C}}\mathbf{V}_h)),\\
        \mathbf{Q}_h=\mathbf{e}_h^{(i)}\mathbf{W}_h^{Q_h}, \mathbf{K}_h=\bm{H}^{(i)}&\mathbf{W}_h^{K_h}, \mathbf{V}_h=\bm{H}^{(i)}\mathbf{W}_h^{V_h},\\
        \mathbf{Q}_o=\mathbf{e}_o^{(i)}\mathbf{W}_o^{Q_o},\mathbf{K}_o=\bm{O}^{(i)}&\mathbf{W}_o^{K_o}, \mathbf{W}_o=\bm{O}^{(i)}\mathbf{W}_o^{V_o},
    \end{aligned}
\end{equation}
where the subscript $h$ and $o$ denote the human branch and the object branch, respectively, $\mathbf{W_h}$ and $\mathbf{W_o}$ denote the trainable parameters of the two branches.

\noindent{\textbf{Losses Formulation.}}
To model the spatial relation between objects, we designed a novel object-pairwise loss.
Our insight is that the relative distance between the interacted objects changes constantly with certain patterns, and explicitly adding constraints on the relative distance between the interacted objects is beneficial.
For example, when cutting an apple with a knife, the knife gradually approaches and stays on the surface of the apple for a while, then moves away. 
We adopt the $\mathbb{L}$2 loss as $\mathcal{L}_{dis}$ to keep the consistency between the generated results and the ground truth.

Besides, we also adopt the widely adopted geometric loss in the field of human motion, including joint position loss $\mathcal{L}_{pos}$ and the joint velocity loss $\mathcal{L}_{vel}$. To ensure the authenticity of our synthesized motion, we additionally utilized the interpenetration loss $\mathcal{L}_{pen}$ between human and objects. Here, we summarize our ultimate optimization objective as:
\begin{equation}
    \mathcal{L}=\lambda_{vel}\mathcal{L}_{vel}+\lambda_{pos}\mathcal{L}_{pos}+\lambda_{pen}\mathcal{L}_{pen}+\lambda_{dis}\mathcal{L}_{dis},
\end{equation}
and we empirically set $\lambda_{vel}=\lambda_{pos}=\lambda_{pen}=1$, and $\lambda_{dis}=0.1$. More details of the formulations of the losses will be illustrated in the supplementary materials.

\subsection{The \texttt{HIMO-SegGen} Framework}
\label{sec:himo_seggen}

\noindent{\textbf{Problem Formulation.}}
The only difference with \texttt{HIMO-Gen} lies in the input text prompts, where \texttt{HIMO-Gen} takes one single text description $\bm{L}$ as a condition, yet \texttt{HIMO-SegGen} is conditioned on a series of several atomic text prompts as $\bm{L}=\{l_1, l_2, \cdots, l_m\}$. 
The key of \texttt{HIMO-SegGen} is to ensure smooth and realistic transitions between the generated HOI clips.

\begin{figure}[tb]
  \centering
  \includegraphics[width=1\textwidth]{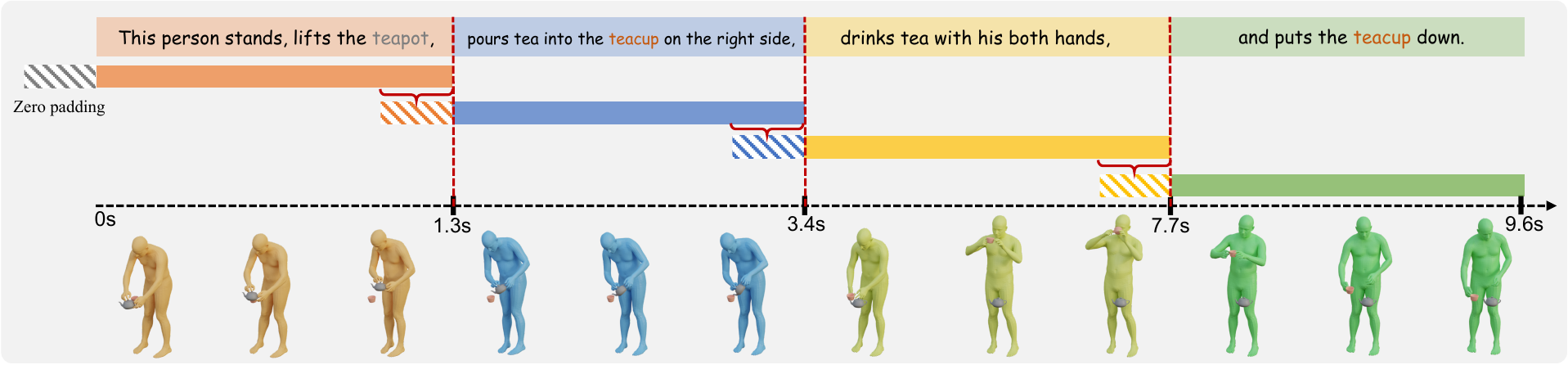}
  \vspace{-0.2cm}
  \caption{\textbf{The auto-regressive generation pipeline.} The auto-regressive pipeline to iteratively generate the HOI synthesis results. We obtain the subsequent motion conditioned on the last few frames of the previously generated motion.}
  \label{fig:test_process}
  \vspace{-0.2cm}
\end{figure}

\noindent{\textbf{Generation Pipeline.}}
Our generation pipeline is demonstrated in~\cref{fig:test_process}, which is simple yet effective.
We segment the long HOI sequences into smaller motion-text pairs so that the natural transitions of consecutive HOIs are reserved. We first train the \texttt{HIMO-Gen} model to empower the ability to generate more fine-granularity HOIs.
Different from the vanilla \texttt{HIMO-Gen} model which is only conditioned on the initial state of the first frame, we modify it to conditioned generation on the past \textit{few} frames. During the inference time, we take the last few frames of the previous generation result as the condition for the next generation, to iteratively obtain the composite HOI results.




\begin{table*}[t]
  \caption{\textbf{Quantitative baseline comparisons} on the \textbf{2-objects} partition of HIMO. $\pm$ indicates the 95\% confidence interval and $\rightarrow$ means the closer the better. \textbf{Bold} indicates best result and \underline{underline} indicates second best.}
  \label{tab:2objects}
  \vspace{-0.2cm}
  \begin{adjustbox}{width=\columnwidth, center}
  \begin{tabular}{@{}lccccc@{}}
    \toprule
    Method & R-Precision(Top 3)$\uparrow$ & FID $\downarrow$ & MM-Dist$\downarrow$ & Diversity$\rightarrow$ &MModality$\uparrow$\\ 
    \midrule
      Real & $0.7988^{\pm0.0081}$ & $0.0176^{\pm0.0065}$ & $3.5659^{\pm0.0109}$ & $11.3973^{\pm0.2577}$& $-$  \\
      \midrule
      IMoS~\cite{ghosh2022imos} & $0.5013^{\pm0.0120}$ & $7.5890^{\pm0.1121}$ & $8.7402^{\pm0.0310}$ & $7.0033^{\pm0.3205}$ & $0.9920^{\pm0.2004}$ \\
      MDM~\cite{tevet2023mdm}& $0.6052^{\pm0.0099}$ & $6.8457^{\pm0.3315}$ & $8.0187^{\pm0.0500}$ & $\mathbf{11.3891^{\pm0.2342}}$ & $1.2880^{\pm0.2110}$ \\
      priorMDM~\cite{priormdm}& $0.5891^{\pm0.0031}$ & $7.8517^{\pm0.2516}$ & $7.2509^{\pm0.0065}$ & $12.5799^{\pm0.1460}$ & $1.5911^{\pm0.1449}$ \\
      \cellcolor{Gray}HIMO-Gen & \cellcolor{Gray} $\underline{0.6369^{\pm0.0032}}$& \cellcolor{Gray} $\mathbf{1.4811^{\pm0.0427}}$& \cellcolor{Gray} $\underline{3.6491^{\pm0.0101}}$ & \cellcolor{Gray} $\underline{11.6603^{\pm0.2043}}$ & \cellcolor{Gray} $\underline{1.7863^{\pm0.0570}}$\\
      \cellcolor{Gray2}HIMO-SegGen  & \cellcolor{Gray2} $\mathbf{0.6404^{\pm0.0449}}$ & \cellcolor{Gray2} $\underline{4.2004^{\pm0.4729}}$  & \cellcolor{Gray2}  $\mathbf{3.6077^{\pm0.0230}}$ & \cellcolor{Gray2} $15.7317^{\pm0.4090}$& \cellcolor{Gray2} $\mathbf{2.0495^{\pm0.0030}}$\\
    \bottomrule
  \end{tabular}
  \end{adjustbox}
\end{table*}

\begin{table*}[t]
  \caption{\textbf{Quantitative baseline comparisons} on the \textbf{3-objects} partition of HIMO.}
  \label{tab:3objects}
  \vspace{-0.2cm}
  \begin{adjustbox}{width=\columnwidth, center}
  \begin{tabular}{@{}lccccc@{}}
    \toprule
    Method & R-Precision(Top 3)$\uparrow$ & FID $\downarrow$ & MM-Dist$\downarrow$ & Diversity$\rightarrow$ &MModality$\uparrow$\\ 
    \midrule
      Real & $0.6988^{\pm0.0054}$ & $0.1811^{\pm0.0442}$ & $3.7696^{\pm0.0279}$ & $9.7674^{\pm0.2180}$ & $-$ \\
      \midrule
      IMoS~\cite{ghosh2022imos}& $0.4662^{\pm0.1010}$ & $4.9902^{\pm0.1772}$ & $7.7702^{\pm0.0585}$ & $\underline{9.2310^{\pm0.1130}}$ & $0.6254^{\pm0.1102}$\\
      MDM~\cite{tevet2023mdm}& $0.5025^{\pm0.0136}$& $\mathbf{4.5713^{\pm0.1100}}$ & $6.3144^{\pm0.0268}$ & $8.8953^{\pm0.2851}$ & $\underline{0.7893^{\pm0.5715}}$\\
      priorMDM~\cite{priormdm}& $\underline{0.5137^{\pm0.0257}}$ & $4.8210^{\pm0.2030}$ & $\underline{5.8900^{\pm0.0232}}$ & $\mathbf{9.3402^{\pm0.0230}}$ & $\mathbf{0.8120^{\pm0.3280}}$\\
      \cellcolor{Gray}HIMO-Gen & \cellcolor{Gray}$\mathbf{0.5350^{\pm0.0185}}$ & \cellcolor{Gray} $\underline{4.7712^{\pm0.1638}}$& \cellcolor{Gray}$\mathbf{5.0866^{\pm0.0412}}$ & \cellcolor{Gray}$8.9460^{\pm0.1378}$ & \cellcolor{Gray}$0.7561^{\pm0.0546}$\\
    \bottomrule
  \end{tabular}
  \end{adjustbox}
\end{table*}

\section{Experiments}

In this section, we elaborate on the dataset and implementation details, the baseline methods and the evaluation metrics. Then we present extensive experimental results with ablation studies to show the effectiveness of our benchmark.

\noindent{\textbf{Dataset Details.}} 
We follow~\cite{Guo_2022_CVPR_humanml} to split our dataset into train, test, and validation sets with the ratio of 0.8, 0.15, and 0.05. 
The maximum motion length is set to 300 for \texttt{HIMO-Gen} and 100 for \texttt{HIMO-SegGen}. The maximum text length is set to 40 and 15, respectively.
To mitigate the influence of the order of several objects, we randomly shuffle the objects when loading the data.

\noindent{\textbf{Implementation Details.}} 
For both \texttt{HIMO-Gen} and \texttt{HIMO-SegGen}, the denoising network consists of $\ell_{enc}=8$ layers of mutual interaction modules with 4 attention heads. We train the network using Adam~\cite{kingma2017adam} optimizer with a learning rate of 0.0001 and weight decay of 0.99. The training of \texttt{HIMO-Gen} and \texttt{HIMO-SegGen} takes about 9 and 12 hours on a single A100 GPU with a batch size of 128.

\noindent{\textbf{Baseline Methods.}}
We adopt the prominent text-to-motion methods MDM~\cite{tevet2023mdm}, priorMDM~\cite{priormdm} and recent intent-driven HOI synthesis method IMoS~\cite{ghosh2022imos} as our baselines. To be specific, we re-implement each model to support the condition input of object geometry and the initial states of human and objects. More details of the baselines are provided in the supplementary materials.

\noindent{\textbf{Evaluation Metrics.}}
Following~\cite{Guo_2022_CVPR_humanml}, we train a motion feature extractor and a text feature extractor in a contrastive manner for evaluation. Since we are evaluating the generation quality of HOIs, the motion features here include both human and object motion. Then we evaluate over the set of metrics proposed by~\cite{Guo_2022_CVPR_humanml}: \textit{R-precision} and \textit{MM-Dist} determine the relevance between the generated motion and the input prompts, \textit{FID} measures the dissimilarity of the generated and ground-truth motion in latent feature space, \textit{Diversity} measures the variability of the generated motions, and \textit{MultiModality} shows the mean-variance of generated motion conditioned on a single text prompt.

\begin{figure}[tb]
  \centering
  \includegraphics[width=1\textwidth]{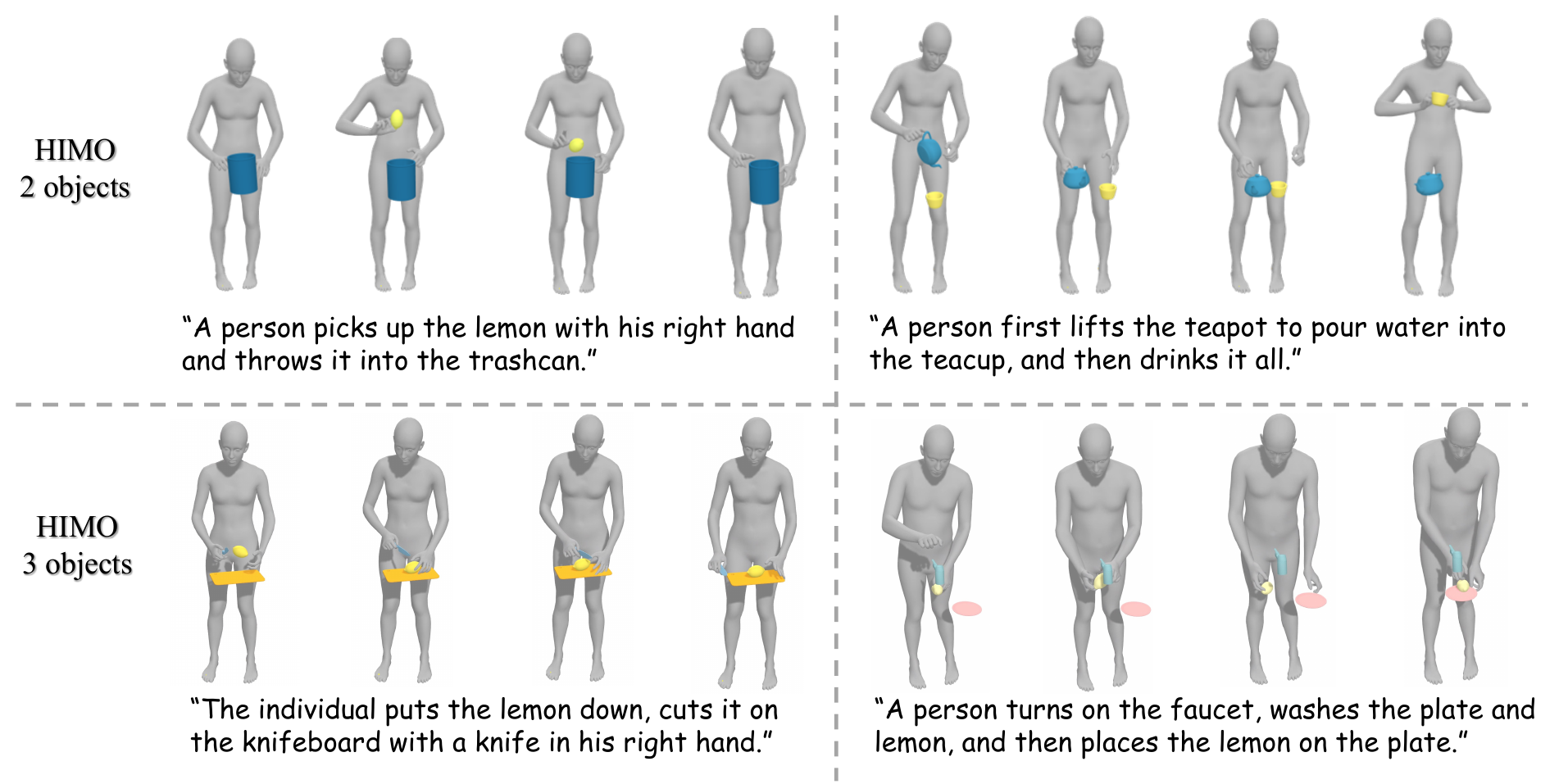}
  \vspace{-0.2cm}
  \caption{\textbf{Visualization results.} Our HIMO-Gen framework can generate plausible and realistic sequences of human interacting with multiple objects.}
  \vspace{-0.2cm}
  \label{fig:vis}
\end{figure}

\subsection{Results and Analysis}

\noindent{\textbf{Quantitative Comparisons.}}
Quantitative results on the 2-objects and 3-objects partitions of HIMO are demonstrated in ~\cref{tab:2objects} and ~\cref{tab:3objects}. In the 2-objects setting, our method HIMO-Gen outperforms the baseline methods on all metrics, which indicates our method has better generation quality over baseline methods. The quantitative results of HIMO-SegGen outperform HIMO-Gen in MM-Dist and MModality, which may be attributed to its fine-grained control of generated motion since HIMO-SegGen takes more concise descriptions as input conditions. 
In the 3-objects setting of HIMO, our method also outperforms baselines on the R-precision and MM-Dist metrics, which further shows the effectiveness of our method.

\noindent{\textbf{Qualitative Results.}}
Qualitative results of our method are depicted in~\cref{fig:vis}, which show that our method can synthesize plausible human-object interaction sequences. More generation results of \texttt{HIMO-SegGen} and visual comparisons will be presented in the supplementary materials.

\noindent{\textbf{Generalization Experiments.}} We find that our trained model can also apply to unseen object meshes. Besides, with our elaborate annotate temporal segments, our trained \texttt{HIMO-SegGen} model can generate \textit{novel} HOI compositions. Due to the space limit, more results are listed in the supplementary materials.

\subsection{Ablation Study}

We conduct extensive ablation studies on \texttt{HIMO-Gen} and \texttt{HIMO-SegGen}, as depicted in ~\cref{tab:ablation} and ~\cref{tab:ablation_seg}. 
For \texttt{HIMO-Gen}, we experiment with three settings: 1) Replacing the \textbf{mutual interaction module} with the original transformer layer (w/o IM); 2) Removing the \textbf{object-pairwise loss} (w/o $\mathcal{L}_{dis}$); 3) Generating human motion conditioned on the generated objects motion as~\cite{li2023objectmotionguided}. From~\cref{tab:ablation}, we can derive that the removal of the mutual interaction module results in a huge drop of R-precision, which verifies the effectiveness of the feature fusion between human and objects. The object-pairwise loss helps a lot in constraining the generated motions. Consecutive generation scheme brings huge performance drop to generate the human and object motions simultaneously.
For \texttt{HIMO-SegGen}, we experiment on the conditioned past 1, 5, 10 and 20 frames. Given the results in~\cref{tab:ablation_seg}, ``10-frames'' achieves the best performance in R-precision, FID and \textit{MM-Dist}. More past frames (``20 frames'') provide no additional performance gain, which may result from less attention to the predicted frames.

\begin{table*}[t]
  \caption{\textbf{Ablation studies} of several model designs on \textbf{2-objects} partition of HIMO.}
  \label{tab:ablation}
  \vspace{-0.2cm}
  \begin{adjustbox}{width=\columnwidth, center}
  \begin{tabular}{@{}lccccc@{}}
    \toprule
    Method & R-Precision(Top 3)$\uparrow$ & FID $\downarrow$ & MM-Dist$\downarrow$ & Diversity$\rightarrow$ & MModality$\uparrow$\\ 
    \midrule
    Real & $0.7988^{\pm0.0081}$ & $0.0176^{\pm0.0065}$ & $3.5659^{\pm0.0109}$ & $11.3973^{\pm0.2577}$& $-$  \\
    \midrule
      Ours-w/o IM& $0.4710^{\pm0.0509}$& $3.5113^{\pm0.0029}$ & $6.7135^{\pm0.1029}$ & $13.0256^{\pm0.1207}$ & $1.2086^{\pm0.0492}$ \\
      Ours-w/o $\mathcal{L}_{dis}$& $\mathbf{0.6426^{\pm0.0085}}$ & $\underline{2.8694^{\pm0.0488}}$ & $\underline{4.9342^{\pm0.0191}}$ & $\underline{12.0789^{\pm0.1153}}$ & $\underline{1.4394^{\pm0.0549}}$ \\
      Ours-consec.& $0.3902^{\pm0.3046}$ & $8.5024^{\pm0.2310}$ & $8.9982^{\pm0.0120}$ & $10.2788^{\pm0.1113}$ & $1.1209^{\pm0.0890}$ \\
      \midrule
      \cellcolor{Gray}HIMO-Gen & \cellcolor{Gray} $\underline{0.6369^{\pm0.0032}}$& \cellcolor{Gray} $\mathbf{1.4811^{\pm0.0427}}$& \cellcolor{Gray} $\mathbf{3.6491^{\pm0.0101}}$ & \cellcolor{Gray} $\mathbf{11.6603^{\pm0.2043}}$ & \cellcolor{Gray} $\mathbf{1.7863^{\pm0.0570}}$\\
    \bottomrule
  \end{tabular}
  \end{adjustbox}
\end{table*}
\begin{table*}[t]
  \caption{\textbf{Ablation studies} of the frame number of conditioned frames for \texttt{HIMO-SegGen} on \textbf{2-objects} partition of the HIMO dataset.}
  \label{tab:ablation_seg}
  \vspace{-0.2cm}
  \begin{adjustbox}{width=\columnwidth, center}
  \begin{tabular}{@{}cccccc@{}}
    \toprule
    \#Frame & R-Precision(Top 3)$\uparrow$ & FID $\downarrow$ & MM-Dist$\downarrow$ & Diversity$\rightarrow$ &MModality$\uparrow$\\ 
    \midrule
    Real & $0.7988^{\pm0.0081}$ & $0.0176^{\pm0.0065}$ & $3.5659^{\pm0.0109}$ & $11.3973^{\pm0.2577}$& $-$  \\
    \midrule
      1 & $0.5502^{\pm0.0230}$ & $8.7320^{\pm0.3208}$ & $6.2094^{\pm0.3553}$ & $\mathbf{13.2999^{\pm0.5034}}$ & $0.8814^{\pm0.0394}$\\
      5 & $0.5448^{\pm0.0493}$& $7.3029^{\pm0.2677}$ & $\underline{5.5053^{\pm0.0026}}$ & $17.8045^{\pm0.0956}$ & $0.9500^{\pm0.2374}$\\
      \cellcolor{Gray}10  & \cellcolor{Gray} $\mathbf{0.6404^{\pm0.0449}}$ & \cellcolor{Gray} $\mathbf{4.2004^{\pm0.4729}}$  & \cellcolor{Gray}  $\mathbf{3.6077^{\pm0.0230}}$ & \cellcolor{Gray} $15.7317^{\pm0.4090}$& \cellcolor{Gray} $\underline{2.0495^{\pm0.0030}}$\\
      20 & $\underline{0.5508^{\pm0.0891}}$& $\underline{5.6749^{\pm0.1002}}$ & $6.7708^{\pm0.0355}$ & $\underline{15.2112^{\pm0.2210}}$ & $\mathbf{2.3863^{\pm0.3399}}$\\
    \bottomrule
  \end{tabular}
  \end{adjustbox}
\end{table*}

\vspace{-0.4cm}
\section{Conclusion}
In this paper, we propose HIMO, a dataset of full-body human interacting with multiple household objects. We collect a large amount of 4D HOI sequences with precise and diverse human motions, object motions and mutual contacts. Based on that, we annotate the detailed textual descriptions for each HOI sequence to enable the text-driven HOI synthesis. We also equip it with elaborative temporal segments to decompose the long sequences into several atomic HOI steps, which facilitates the HOI synthesis driven by consecutive text prompts. Extensive experiments show plausible results, verifying that the HIMO dataset is promising for downstream HOI generation tasks.

\noindent\textbf{Limitations:} Our work has the following limitations: 
1) \textbf{Large objects:} Our HIMO dataset is built mainly based on small household objects, without human interacting with large objects such as chairs and suitcases.
2) \textbf{RGB modality:} Though we set up a Kinect camera to capture the monocular RGB sequences, the human and object appearances are unnatural since humans are wearing tight MoCap suits and objects are attached with reflective markers.
3) \textbf{Facial expressions:} We neglect the facial expressions since our dataset mainly focuses on the manipulation of multiple objects, \ieno, the body movements, finger gestures and the object motions, which have little correlation with facial expressions.


\section*{Acknowledgements}
This work was supported in part by NSFC (62201342, 62101325), and Shanghai Municipal Science and Technology Major Project (2021SHZDZX0102).The authors would like to appreciate the High Performance Computing Center at Eastern Institute of Technology, Ningbo for the GPU support.


%
%
\bibliographystyle{splncs04}
\bibliography{main}

\clearpage

\appendix
\begingroup

\appendix
\begin{center}
\Large{\bf HIMO: A New Benchmark for Full-Body Human Interacting with Multiple Objects \\ **Appendix**}
\end{center}

\renewcommand{\thefigure}{\Alph{figure}}
\renewcommand{\thetable}{\Alph{table}}

\setcounter{page}{1}
\setcounter{table}{0}
\setcounter{figure}{0}

\section{Loss Formulations of \texttt{HIMO-Gen}}

In order to train the \texttt{HIMO-Gen} framework, we adopt four types of losses specifically designed for our task of text-driven human-object interaction (HOI) synthesis. 

To generate authentic human motion, we adopt the widely used geometric loss in recent text-to-motion works~\cite{priormdm,tevet2023mdm,chen2023mldm}, including joint position loss $\mathcal{L}_{pos}$ and joint velocity loss $\mathcal{L}_{vel}$. Formally, we denote the ground truth $j$-th joint position of human of the $n$-th frame as $P_n^j$, and the generated result as $\hat{P}_n^j$. The joint position loss $\mathcal{L}_{pos}$ is formulated as:
\begin{equation}
    \mathcal{L}_{pos}=\frac{1}{N}\sum\limits_{n=1}\limits^{N}\sum\limits_{j=1}\limits^{J}\|P_n^j-\hat{P}_n^j\|_2^2,
\end{equation}
where $N$ is the length of human motion and $J$ is the number of joints. Similarly, the joint velocity loss $\mathcal{L}_{vel}$ is formulated as:
\begin{equation}
    \mathcal{L}_{vel}=\frac{1}{N-1}\sum\limits_{n=1}\limits^{N-1}\sum\limits_{j=1}\limits^{J}\|(P_{n+1}^j-P_n^j)-(\hat{P}_{n+1}^j-\hat{P}_n^j)\|_2^2.
\end{equation}

Following~\cite{jiang2020multiperson}, a signed distance field (SDF) based interpenetration loss $\mathcal{L}_{pen}$ is also adopted to prevent the collision between human and objects. Let $\phi$ be a modified SDF of the human body defined as follows:
\begin{equation}
    \phi(x,y,z)=-\min(SDF(x,y,z),0).
\end{equation}
According the definition above, points inside the human body have positive values of $\phi$ proportional to the distance from the surface, otherwise it equals 0 outside of the human body. Here we define $\phi$ on a voxel grid of dimensions $N_h\times N_h\times N_h$, where $N_h=32$. For the object $o$, its $i$-th sampled point on the surface is denoted as $v_o^i$. Thus, the interpenetration loss of object $o$ colliding with the human is defined as :
\begin{equation}
    P_o=\sum\limits_{i=1}\limits^{S}\tilde{\phi}(v_o^i),
\end{equation}
where $S$ is the number of sampled points on the surface, and $\tilde{\phi}(v_o^i)$ samples the $\phi$ value for each 3D point $v_o^i$ in a differentiable way from the 3D grid via trilinear interpolation. Then the interpenetration loss for all objects is formulated as:
\begin{equation}
    \mathcal{L}_{pen}=\sum\limits_{o\in O}P_o,
\end{equation}
where $O$ is the set of the involved objects.

To model the spatial relation between objects, we further use an object-pairwise loss $\mathcal{L}_{dis}$. Formally, we denote the sampled points set of object $i$ and $j$ during the whole HOI sequence as $V_i^{1:N}$ and $V_j^{1:N}$, and the points set transformed by the generated motion as $\hat{V}_i^{1:N}$ and $\hat{V}_j^{1:N}$. We then have the distance between the two objects as $\Delta V_{ij}=\|V_i^{1:N}-V_j^{1:N}\|_2^2$. Our insight is that this distance between them follows certain pattern, so that we adopt an $\mathbb{L}$2 loss to keep the consistency between the generated results and the ground truth. The object-pairwise loss is then formulated as :
\begin{equation}
    \mathcal{L}_{dis}=\sum\limits_{i\neq j}\|\Delta V_{ij}-\Delta\hat{V}_{ij}\|_2^2.
\end{equation}

\section{Implementation Details of Baselines}
We re-implement MDM~\cite{tevet2023mdm}, PriorMDM~\cite{priormdm} and IMoS~\cite{ghosh2022imos} to support the condition input of of object geometries and initial states of human and objects. Below we detailed the implementation of each model.

\noindent{\textbf{MDM.}}~\cite{tevet2023mdm}
We extend the original feature dimensions of the input and output in MDM~\cite{tevet2023mdm} from $D_h$ to $D_h+D_o$, where $D_h$ denotes the dimension of human motion representation and $D_o$ denotes that of object motion representation. To embed the condition input of object geometry, we feed it into a linear layer and concatenate it with initial pose of the object. Then all conditions are concatenated with the noised input into the motion embedding.

\noindent{\textbf{PriorMDM.}}~\cite{priormdm}
The original PriorMDM~\cite{priormdm} is intended for dual-person motion generation with two branches of MDM and one singular ComMDM to coordinate the two branches. We modify the two human-motion branches into a human-motion branch and an object-motion branch. Also we place the ComMDM module after the 4-th transformer layer of each branch to enable the communication between the two branches. 

\noindent{\textbf{IMoS.}}~\cite{ghosh2022imos}
IMoS~\cite{ghosh2022imos} is an intent-driven HOI synthesis model with the architecture of VAE~\cite{kingma2022autoencoding}. We modify the input action label into our text prompt and integrate the arm and body synthesis module into one. Additionally, the movement of objects is also generated through the VAE instead of the original optimization module. 

\section{Object Categories and Interaction Settings}
\subsection{Object Categories}
\label{subsec:oc}
We list all 53 daily-life objects of our HIMO dataset in~\cref{tab:obj_cat}.

\begin{table*}[t]
  \caption{\textbf{The object categories of the HIMO dataset.}}
  \label{tab:obj_cat}
  \vspace{-0.2cm}
  \begin{adjustbox}{width=\columnwidth, center}
  \begin{tabular}{|l|l|l|l|l|l|l|}
    \toprule
    01. Plate&02. Pan&03. Teapot&04. Washbasin&05. Bowl&06. Trashcan&07. Dustpan \\ \midrule
    08. Flowerpot&09. Phoneholder &10. Glasses case &11. Knife & 12. Spatula & 13. Beer & 14. Spoon \\ \midrule
    15. Knife board&16. Broom &17. Sprinkler &18. Toothbrush &19. Toothpaste & 20. Faucet & 21. Phone \\ \midrule
    22. Bulb&23. Lampholder &24. Television &25. Remote controller & 26. Hammer & 27. Notebook& 28. Pen \\ \midrule
    29. Glasses& 30. Camera & 31. Laptop & 32. Mouse & 33. Headset &34. Cube-M&35. Cube-L  \\ \midrule
    36. Cylinder-M &37. Cylinder-L &38. Pyramid-M & 39. Pyramid-L &40. Banana &41. Apple &42. Lemon \\ \midrule
    43. Cabbage &44. Pepper &45. Teacup &46. Mug &47. Goblet &48. Bottle &49. Plane \\ \midrule
    50. Piggybank &51. Rubber duck &52. Stanford rabbit & 53. Train & & &\\ 
    \bottomrule
  \end{tabular}
  \end{adjustbox}
\end{table*}

\subsection{Interaction Settings}
We list our interaction settings in~\cref{tab:inter_setting}. Here A, B and C denote objects in~\cref{subsec:oc}.
\begin{table*}[t]
  \caption{\textbf{The interaction settings of the HIMO dataset.}}
  \label{tab:inter_setting}
  \vspace{-0.2cm}
  \begin{adjustbox}{width=\columnwidth, center}
  \begin{tabular}{|l|l|l|}
    \toprule
    01. Put A(and B) into C &
    02. Throw A(and B) into C &
    03. Wash A(and B) under faucet \\ \midrule
    04. Cut A on the knife board &
    05. Pour water into A &
    06. Transfer water between A and B \\ \midrule
    07. Hit A with a hammer &
    08. Play A with B &
    09. Cook \\ \midrule
    10. Have a meal &
    11. Sweep the table &
    12. Brush teeth \\ \midrule
    13. Place A on B &
    14. Use A and B &
    15. Stack A(and B) on C \\ 
    \bottomrule
  \end{tabular}
  \end{adjustbox}
\end{table*}

\subsection{Object Combinations}
We visualize the distribution of object combinations in ~\cref{fig:network_obj}

\begin{figure}[tb]
  \centering
  \includegraphics[width=1\textwidth]{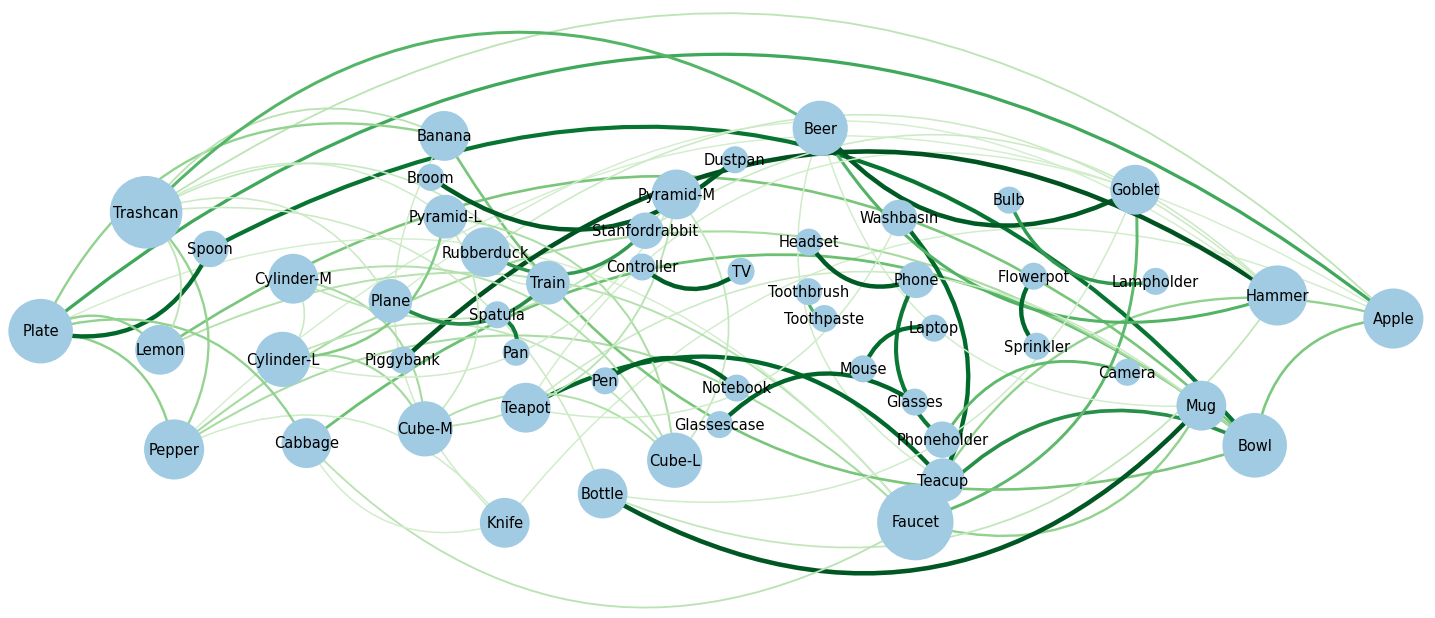}
  \vspace{-0.2cm}
  \caption{\textbf{Visualization of the distribution of object combinations.}}
  \label{fig:network_obj}
  \vspace{-0.2cm}
\end{figure}

\section{Generalization Experiments}
\noindent{\textbf{Unseen object geometries.}}
We experiment on the generalization ability to unseen object geometries of our \texttt{HIMO-Gen} model. We choose several object meshes that have the same category as our dataset but different geometries. We feed their geometries as input conditions to our model and synthesize corresponding human and objects motion. Visualization results are shown in~\cref{fig:unseen_obj}. We can observe that, to some extent, our model can generalize to unseen object meshes despite of some flaws of human-object contact, which may result from the deficiency of their geometric information in our model.

\noindent{\textbf{Novel HOI compositions.}}
We experiment on the ability of our \texttt{HIMO-SegGen} to generate \textit{novel} HOI compositions. We choose several HOI combinations that never appear in our training set, for instance, ``Knocks on the beer with a hammer'' $\to$ ``Grabs the beer to observe'' $\to$ ``Puts them down on the table''. Then we feed the texts to our \texttt{HIMO-SegGen} consecutively to auto-regressively generate HOI clips. Visualization results are depicted in~\cref{fig:hoi_comp}.

\begin{figure}[tb]
  \centering
  \includegraphics[width=1\textwidth]{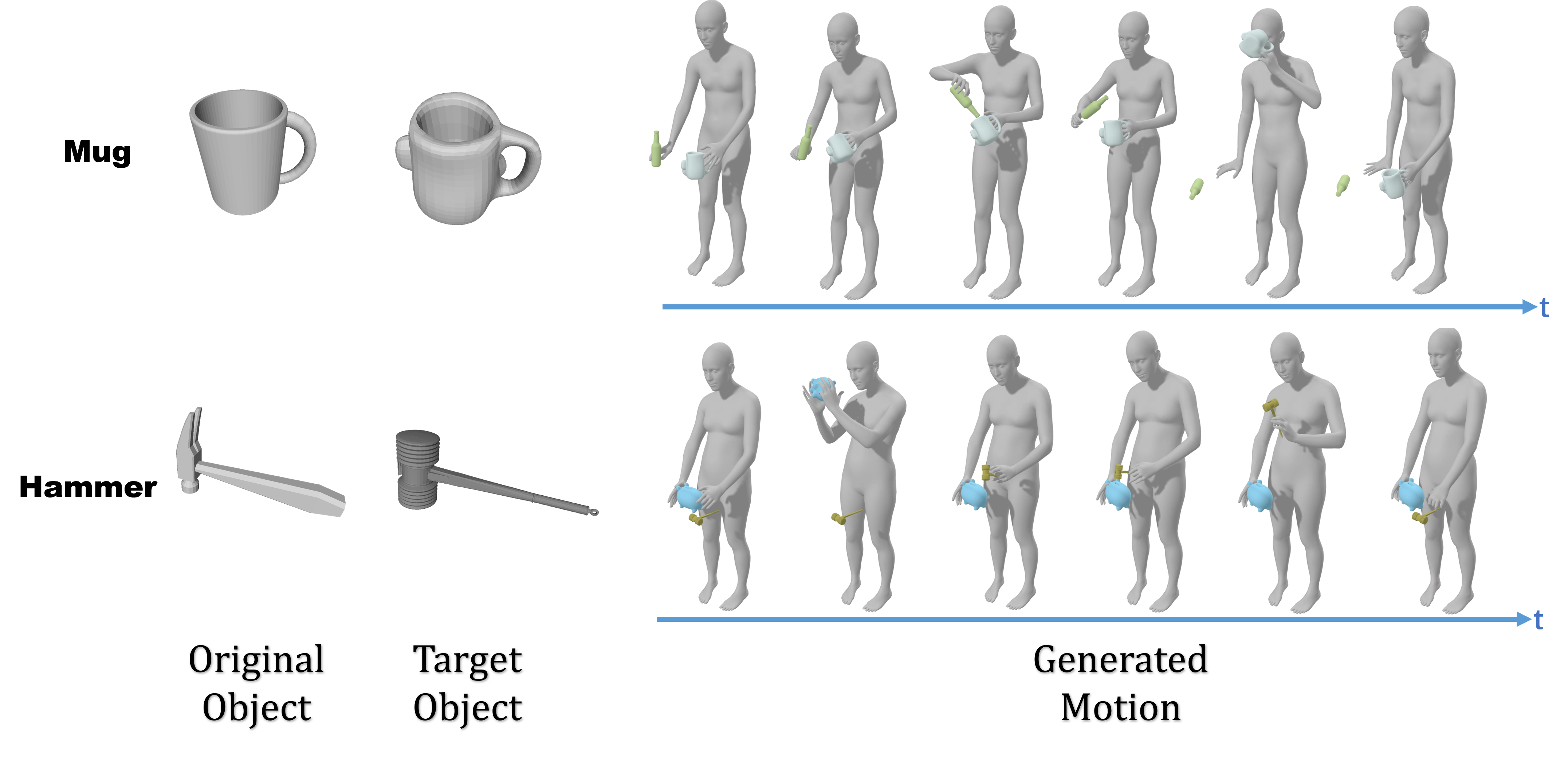}
  \vspace{-0.2cm}
  \caption{\textbf{Generalization experiment on unseen object meshes.}}
  \label{fig:unseen_obj}
  \vspace{-0.2cm}
\end{figure}

\begin{figure}[tb]
  \centering
  \includegraphics[width=1\textwidth]{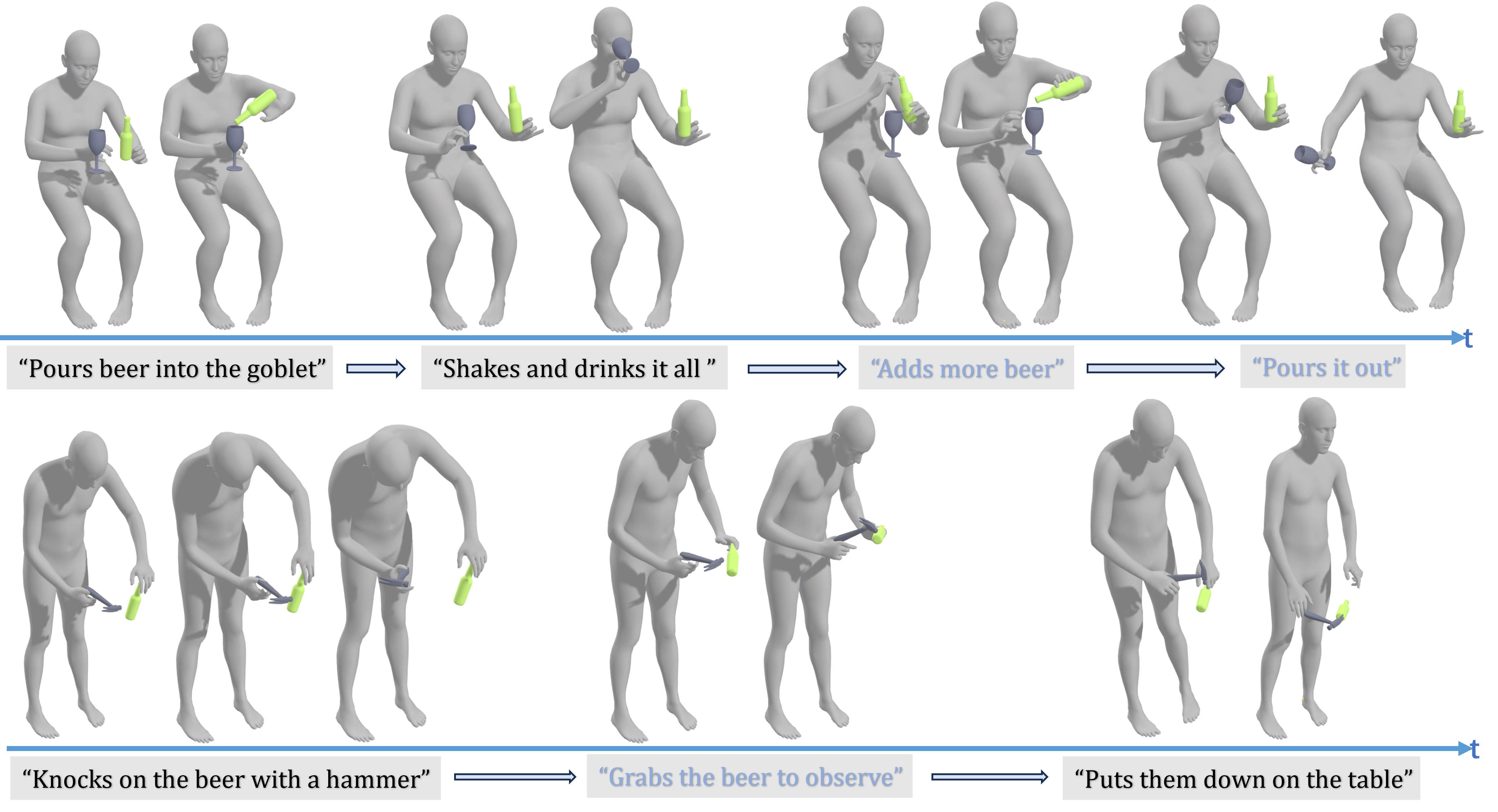}
  \vspace{-0.2cm}
  \caption{\textbf{Generalization experiment on novel HOI compositions.} \textcolor[RGB]{157,195,230}{The text in blue} denotes the \textit{novel} HOI action.}
  \label{fig:hoi_comp}
  \vspace{-0.2cm}
\end{figure}

\section{More Visualization Results}
\noindent{\textbf{Visualization of Dataset.}}
We present more samples from our HIMO dataset in~\cref{fig:data_vis_1} and~\cref{fig:data_vis_2}. Note that we segment both the motion and text descriptions into several clips. Additional visualization results are presented in the supplementary video.

\section{Societal impacts and responsibility to human subjects.}
Our dataset can be leveraged to generate plausible human interactions with multiple objects, which may lead to the creation and spread of misinformation. 
For privacy concerns, all performers signed the agreement on the release of their motion data for research purposes. 
We focus on the pure motion rather than RGB, thus no RGB videos are released and their identity will not be leaked.

\noindent{\textbf{Visualization of HIMO-SegGen.}}
The generated results of \texttt{HIMO-SegGen} are shown in~\cref{fig:seg_vis}. Each text prompt is sent to \texttt{HIMO-SegGen} and generates motion clips auto-regressively conditioned on the last 10 frames of the previous clip.

\noindent{\textbf{Visualization of HIMO-Gen.}}
We show qualitative comparisons of our \texttt{HIMO-Gen} methods with baseline methods in~\cref{fig:comp_vis}. From the figure we can see that our methods keep both consistency and semantics over the baseline methods.

\begin{figure}[tb]
  \centering
  \includegraphics[width=1\textwidth]{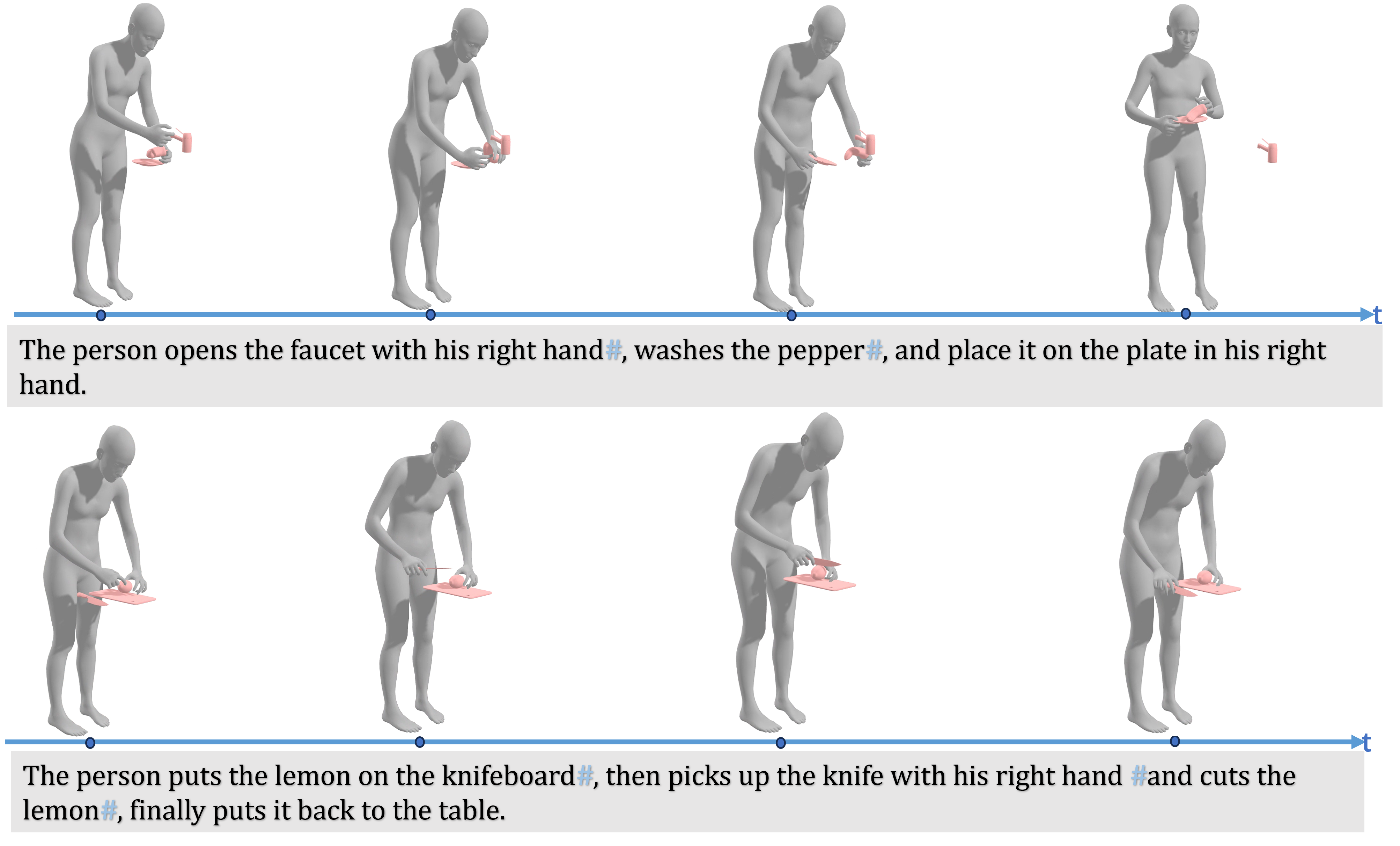}
  \vspace{-0.2cm}
  \caption{\textbf{More visualization samples of our HIMO dataset.} \textcolor[RGB]{157,195,230}{\textbf{\#}} is the separator of our temporal segments.}
  \label{fig:data_vis_1}
  \vspace{-0.2cm}
\end{figure}

\begin{figure}[tb]
  \centering
  \includegraphics[width=1\textwidth]{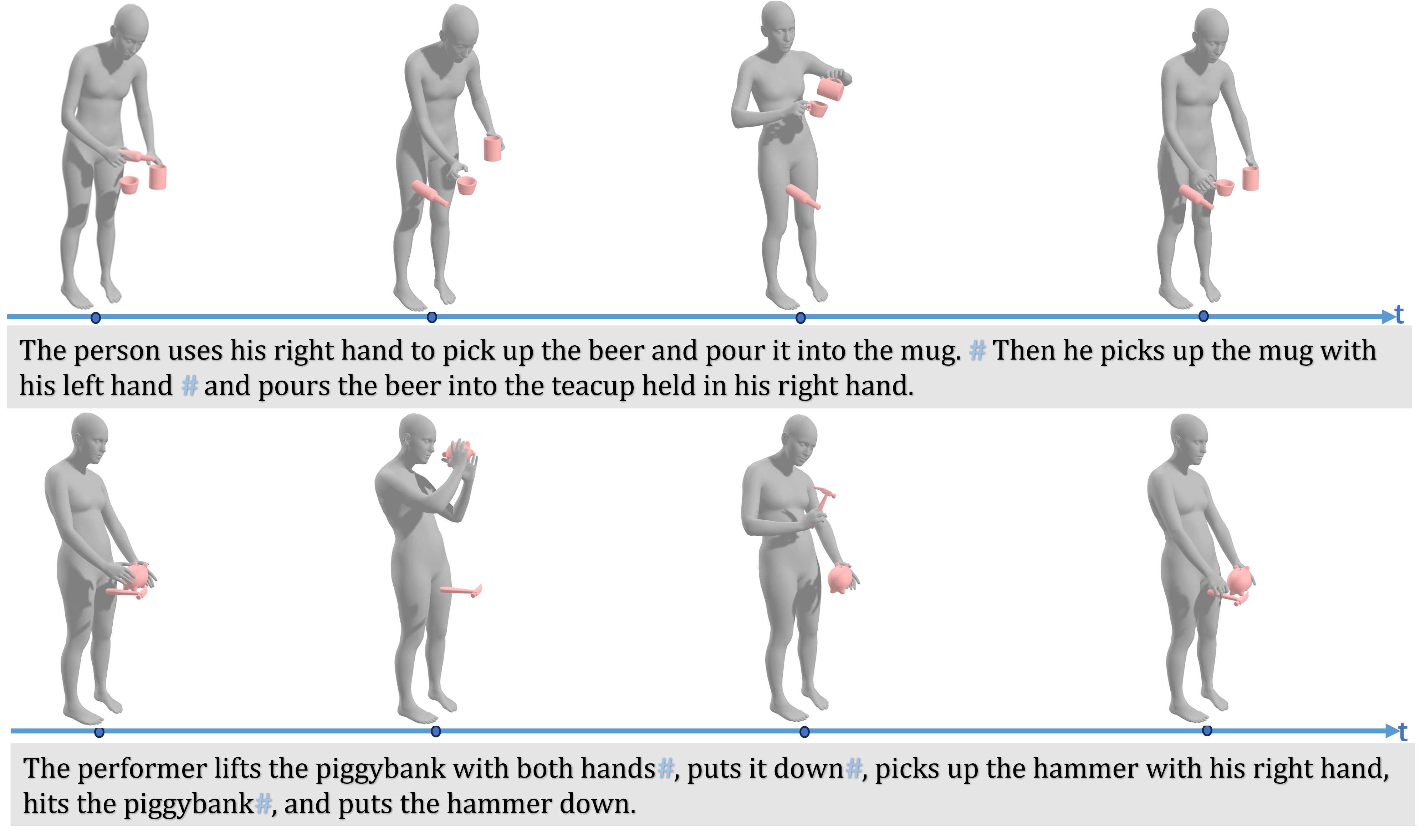}
  \vspace{-0.2cm}
  \caption{\textbf{More visualization samples of our HIMO dataset.} \textcolor[RGB]{157,195,230}{\textbf{\#}} is the separator of our temporal segments. }
  \label{fig:data_vis_2}
  \vspace{-0.2cm}
\end{figure}

\begin{figure}[tb]
  \centering
  \includegraphics[width=1\textwidth]{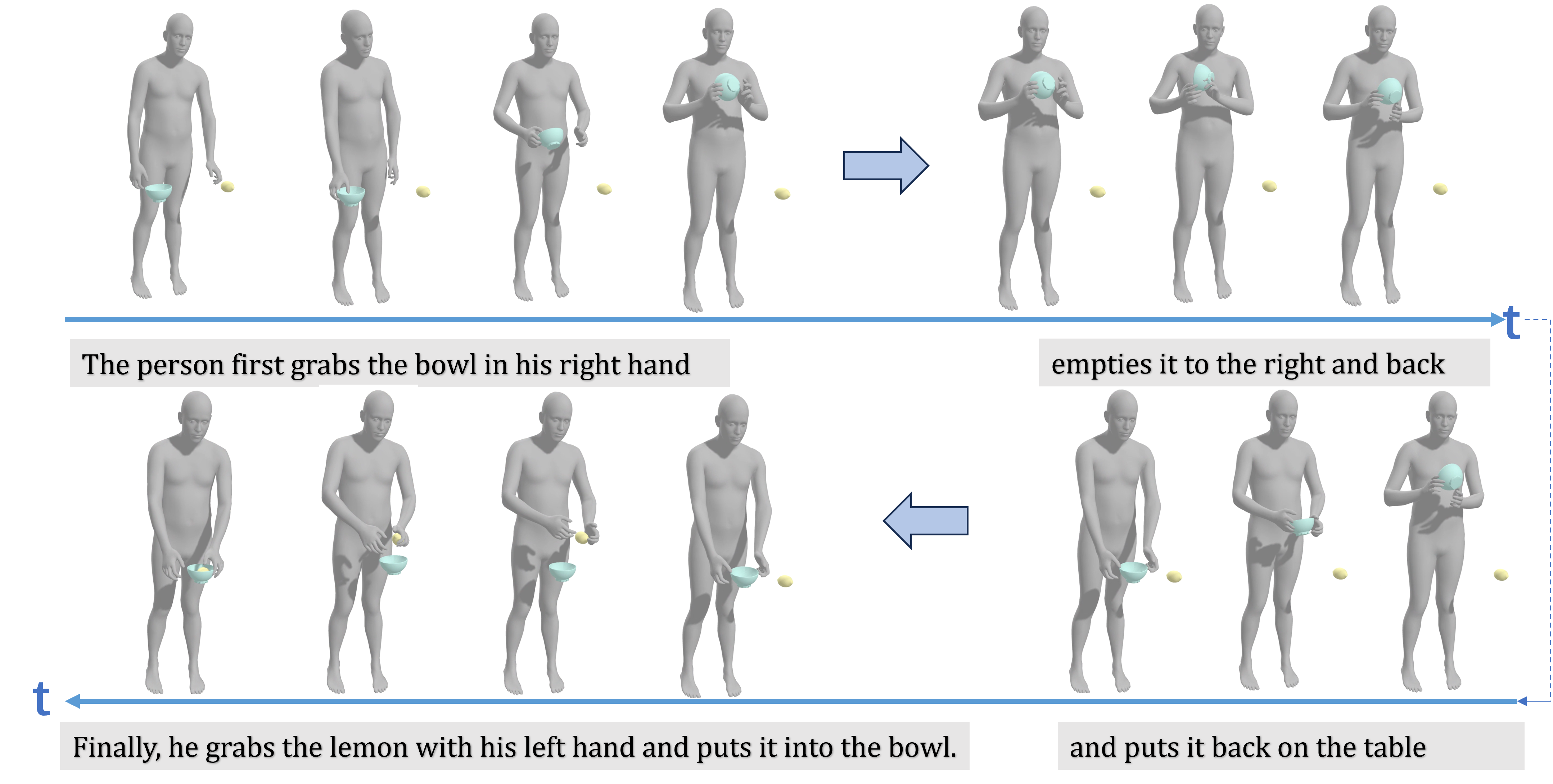}
  \vspace{-0.2cm}
  \caption{\textbf{Visualization results of \texttt{HIMO-SegGen}}.}
  \label{fig:seg_vis}
  \vspace{-0.2cm}
\end{figure}
\clearpage
\begin{figure}[tb]
  \centering
  \includegraphics[width=1\textwidth]{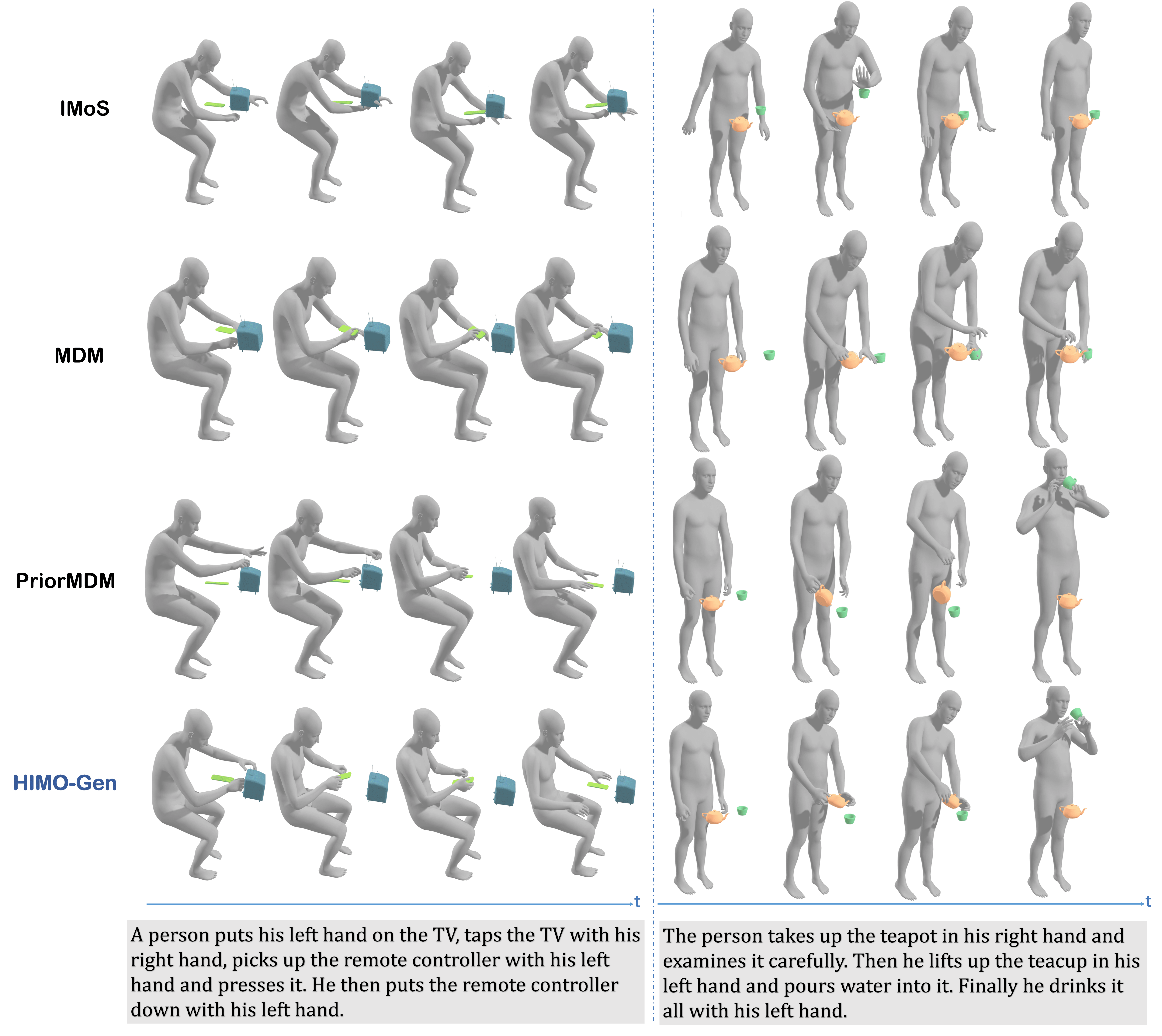}
  \vspace{-0.2cm}
  \caption{\textbf{Visualization comparisons of \texttt{HIMO-Gen} and baselines}.}
  \label{fig:comp_vis}
  \vspace{-0.2cm}
\end{figure}

\clearpage
%
%


\end{document}